\documentclass{article} 
\usepackage{iclr2027_conference,times}
\usepackage{amsmath,amsfonts}

\usepackage{url}
\usepackage{xurl}
\usepackage{graphicx}
\usepackage{booktabs}
\usepackage{colortbl}
\usepackage{wrapfig}
\usepackage[ruled,vlined,linesnumbered]{algorithm2e}
\usepackage{hyperref}
\usepackage{placeins}
\usepackage{tcolorbox}

\newcommand{\experimentplot}[3]{%
  \IfFileExists{#1}{%
    \includegraphics[width=\linewidth,height=#2,keepaspectratio]{#1}%
  }{%
    \fbox{\parbox[c][#2][c]{\dimexpr\linewidth-2\fboxsep-2\fboxrule\relax}{%
      \centering\small #3\par\medskip
      \footnotesize\texttt{\detokenize{#1}}}}%
  }%
}

\title{BACAM: Behavior-Aware Continual Agent Merging for Multi-Turn Interaction}
\author{%
\textbf{Shuaitong Li\textsuperscript{1,2}\thanks{Equal contribution.}, Baochen Xiong\textsuperscript{1,2,3,*}, Xiaoshan Yang\textsuperscript{1,2,3}\thanks{Corresponding author.}, Xizhe Zheng\textsuperscript{4}}\\
\textbf{Yifan Xu\textsuperscript{5}, Jianhao Huang\textsuperscript{1,2,3}, Changsheng Xu\textsuperscript{1,2,3}}\\
\normalfont\textsuperscript{1}State Key Laboratory of Multimodal Artificial Intelligence Systems,\\
\normalfont Institute of Automation, Chinese Academy of Sciences, Beijing, China\\
\normalfont\textsuperscript{2}School of Artificial Intelligence, University of Chinese Academy of Sciences, Beijing, China\\
\normalfont\textsuperscript{3}Pengcheng Laboratory, Shenzhen, China\\
\normalfont\textsuperscript{4}Institute of Computing Technology, Chinese Academy of Sciences\\
\normalfont\textsuperscript{5}King Abdullah University of Science and Technology, Thuwal, Saudi Arabia\\
\normalfont\texttt{lishuaitong2025@ia.ac.cn, xiongbaochen2022@ia.ac.cn}\\
\normalfont\texttt{xiaoshan.yang@nlpr.ia.ac.cn}
}

\iclrfinalcopy
\begin{document}

\maketitle
\lhead{BACAM: Behavior-Aware Continual Agent Merging for Multi-Turn Interaction}

\begin{abstract}
Model merging offers a way to integrate the capabilities of specialized experts, but existing agent merging methods typically require all of them to be available at once. We study continual agent merging, which integrates incoming experts sequentially without retaining previously merged experts. Yet merging in parameter space or feature subspaces does not ensure that the merged model acquires an incoming expert's behavior on interaction trajectories. Moreover, updates toward a new expert can disrupt the merged model's previously integrated interactive behavior. Therefore, we propose Behavior-Aware Continual Agent Merging (BACAM), which learns parameter-wise merging gates from candidate-generated trajectories using expert-guided behavioral supervision. Task-level stability--plasticity control and tensor-level conflict-aware update budgets limit interference with existing capabilities while allowing new ones to be acquired. The learned gates are folded into the model weights without additional inference-time parameters. Across four interactive tasks---web shopping, tool use, information retrieval, and embodied interaction---BACAM achieves an average success rate of 62.82\%, exceeding the strongest evaluated merging baseline by 21.69 percentage points.
\end{abstract}

\section{Introduction}
\label{sec:introduction}

Large language model (LLM) agents can reason and act through multi-turn interactions with tools and environments~\citep{yao2022react,liu2024agentbench}. Recent work has produced specialized experts for tool use, information search, embodied interaction, and web shopping~\citep{qian2026toolrl,chen2025research,feng2026group,shridhar2020alfworld,yao2022webshop}. However, current LLM-based agents often exhibit limited cross-environment robustness~\citep{zeng2024agenttuning}. Joint multi-task RL can train an agent across environments, but requires substantial engineering and computation~\citep{chen2026scaling,luo2025agent,liu2025gem}.

Model merging offers a more efficient approach for models to adapt to diverse environments: it combines specialized models directly in parameter space to support a broader range of tasks, without retraining on a mixture of all tasks~\citep{yang2026model}. Existing methods such as Task Arithmetic, TIES-Merging, DARE, and TSV-Merge have demonstrated the feasibility of integrating multiple task-specific capabilities through parameter merging~\citep{ilharco2022editing,yadav2023ties,yu2024language,gargiulo2025task}.

Model merging has increasingly been extended to LLM-based agents, where current approaches primarily fall into two categories: parameter-space and subspace-based merging. Parameter-space approaches modify model parameters directly, either separating shared and task-specific regions for selective aggregation~\citep{yuan2026behavior} or transferring behavior-critical neurons identified from activation patterns~\citep{feng2026arm}. Subspace-based approaches preserve existing agent behavior by identifying behavior-critical feature subspaces and projecting new task vectors into the corresponding null spaces~\citep{wang2026m2a}. These approaches require jointly available experts for batch merging; repeating the merge as new tasks arise becomes costly~\citep{tang2025merging}.

\begin{wrapfigure}{r}{0.43\linewidth} 
  \includegraphics[width=\linewidth]{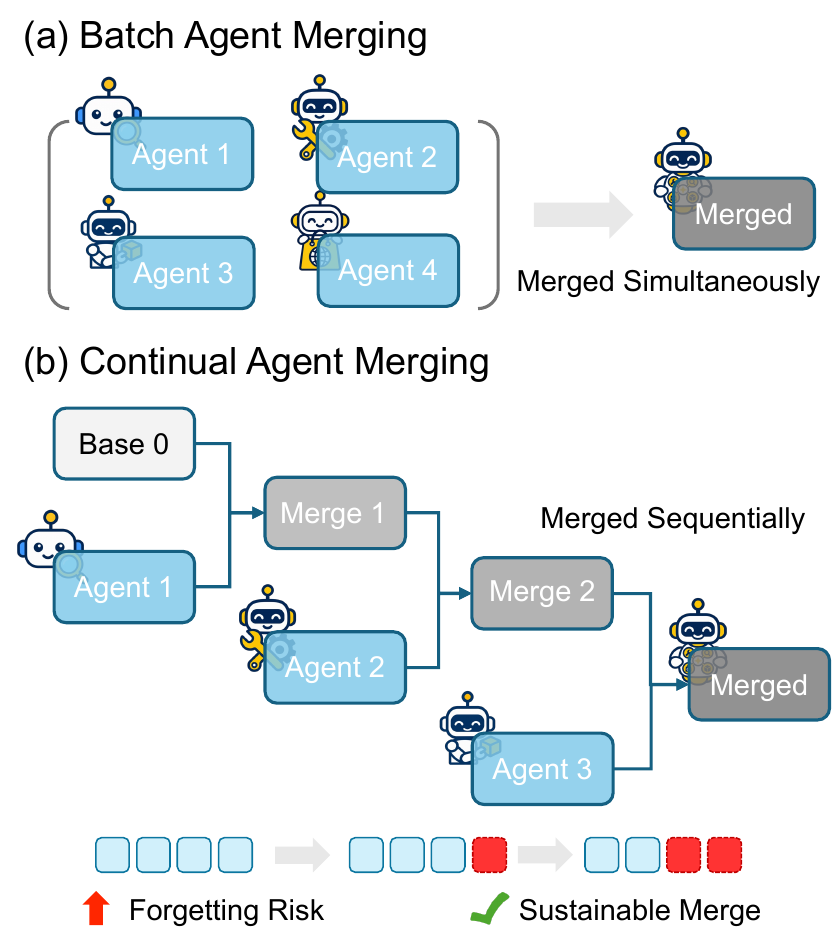}
  \caption{(a) Batch agent merging combines all experts at once. (b) Continual agent merging integrates each incoming expert into the current merged model. Red blocks indicate potential forgetting of earlier capabilities.}
  \label{fig:continual-merging}
\end{wrapfigure}

To this end, we formulate a novel task: continual agent merging for interactive environments. As shown in Figure~\ref{fig:continual-merging}, batch merging combines all experts at once, whereas continual merging updates the current model as new experts arrive without retaining earlier experts.
This continual setting presents two coupled challenges. First, merging in parameter space or feature subspaces does not ensure that the merged model acquires the incoming expert's interactive capability. The merged model's actions determine subsequent observations and may lead to trajectories different from those of the expert. Parameter merging must therefore transfer the expert's behavior to the trajectories induced by the merged model itself. Second, each merge moves the parameters toward the incoming expert, while previously integrated capabilities are encoded in the current merged model. Even a small parameter change can alter an action on an earlier task and redirect the subsequent trajectory, disrupting established interactive behavior. The merge must therefore acquire the incoming expert's capability while preserving the current model's established interactive behavior.

To address these challenges, we propose Behavior-Aware Continual Agent Merging (BACAM), which learns parameter-wise gates between the current merged model and each incoming expert. BACAM comprises two components. First, task-level stability--plasticity control supervises gate updates on candidate-generated interaction trajectories: the incoming expert guides new-task acquisition, while the current merged model constrains behavioral drift on earlier tasks. Second, tensor-level conflict-aware plasticity budgeting assigns smaller update budgets to tensors where these goals conflict, protecting existing capabilities while allowing adaptation in less sensitive regions. Together, these components enable behavior-aware and conflict-sensitive continual agent merging.
We evaluate BACAM across expert counts and merging orders. Under the default order, BACAM has the highest Avg SR at each expert count and exceeds WUDI-Merging by 21.69 percentage points after four experts. Our contributions are:

\begin{enumerate}
\item We formulate continual agent merging for multi-turn interactive tasks and assess capability acquisition and retention over complete interaction trajectories.

\item We introduce BACAM, which learns parameter-wise merging gates from behavioral supervision on candidate-generated trajectories. Task-level constraints guide new-task acquisition while limiting drift on earlier tasks, and tensor-level conflict-aware budgets restrict updates where these goals interfere.

\item Across four interactive tasks, BACAM achieves the highest average success rate among the evaluated merging methods at each stage of the default merging sequence. We also track capability acquisition and retention across four arrival orders and use ablations to measure the contribution of each control component.
\end{enumerate}

\section{Related Work}
\label{sec:related-work}

\subsection{Model Merging}

Model merging combines specialized models in parameter space to integrate their capabilities~\citep{yang2026model}. Task Arithmetic represents fine-tuning updates as task vectors and adds them together~\citep{ilharco2022editing}, but direct addition can cause interference. TIES-Merging trims small updates and resolves sign conflicts~\citep{yadav2023ties}, while DARE sparsifies updates before merging~\citep{yu2024language}. TSV-Merge instead reduces interference by decorrelating task singular vectors~\citep{gargiulo2025task}. Rather than applying predefined rules to task updates, another line of work optimizes the merge directly: WUDI-Merging minimizes interference among task vectors~\citep{cheng2025whoever}, AdaMerging optimizes coefficients using prediction entropy~\citep{yang2024adamerging}, and Expert Merging learns coefficients through unsupervised expert alignment~\citep{zhang2026expert}. These methods improve the combination of model parameters, but do not directly assess whether an agent retains its behavior over multi-turn interactions.

\subsection{Agent Model Merging}

Agent model merging applies these ideas to specialized LLM agents. RAM separates sparse RL task vectors into shared and task-specific regions, then selectively aggregates them to preserve domain-specific behavior~\citep{yuan2026behavior}. ARM identifies role-conditioned neurons from activation patterns and transfers them to a generalist agent~\citep{feng2026arm}. M2A takes a subspace approach: it identifies behavior-critical features and projects new task vectors into their null spaces to limit disruption of existing capabilities~\citep{wang2026m2a}. These methods merge jointly available source models and do not address sequentially arriving experts.

\subsection{Continual Model Merging}

Continual model merging integrates task-specific models sequentially. OPCM projects incoming task vectors onto directions orthogonal to accumulated updates~\citep{tang2025merging}. DOP applies separate projections to protect existing knowledge and acquire new capabilities without task data~\citep{yang2025continual}. MINGLE combines gated low-rank experts for test-time adaptation~\citep{qiu2025mingle}, while NUFILT couples null-space filtering with a lightweight adapter to recover useful new-task directions~\citep{qiu2026null}. These methods primarily evaluate non-interactive prediction tasks. Whether they preserve an agent's earlier capabilities over full interactions remains unclear: one changed action can redirect the subsequent observations and decisions.

\section{Behavior-Aware Continual Agent Merging}
\label{sec:method}

\subsection{Problem Definition}
\label{sec:problem-definition}

We consider an ordered sequence of interactive tasks $\tau=1,\ldots,T$, such as web shopping, tool use, information retrieval, and embodied interaction. Task $\tau$ provides a specialized expert $E_\tau$, an interaction environment, and unlabeled instances $\mathcal{D}_\tau$ without accompanying ground-truth responses or expert demonstration trajectories. All experts share the same architecture and tokenizer. Let $M_t$ be the merged model at stage $t$. We start with $M_1=E_1$. At each subsequent stage $t\geq2$, we merge the current model $M_{t-1}$ with the incoming expert $E_t$ to obtain $M_t$. Earlier experts are no longer retained, but the instances and environments for all integrated tasks remain available for collecting trajectories, following the idea of replaying retained task inputs~\citep{marouf2025ask}. Behavioral supervision is provided by the incoming expert and the current merged model on candidate-generated trajectories.

The goal is for $M_t$ to acquire $E_t$'s capability on the new task while preserving $M_{t-1}$'s capabilities on every earlier task $\tau<t$. We evaluate both acquisition and preservation in multi-turn interactions. Appendix~\ref{sec:preliminaries} gives the formal interaction model.

\subsection{Framework Overview}
\label{sec:framework-overview}

\begin{figure}[t]
  \centering
  \includegraphics[width=\linewidth]{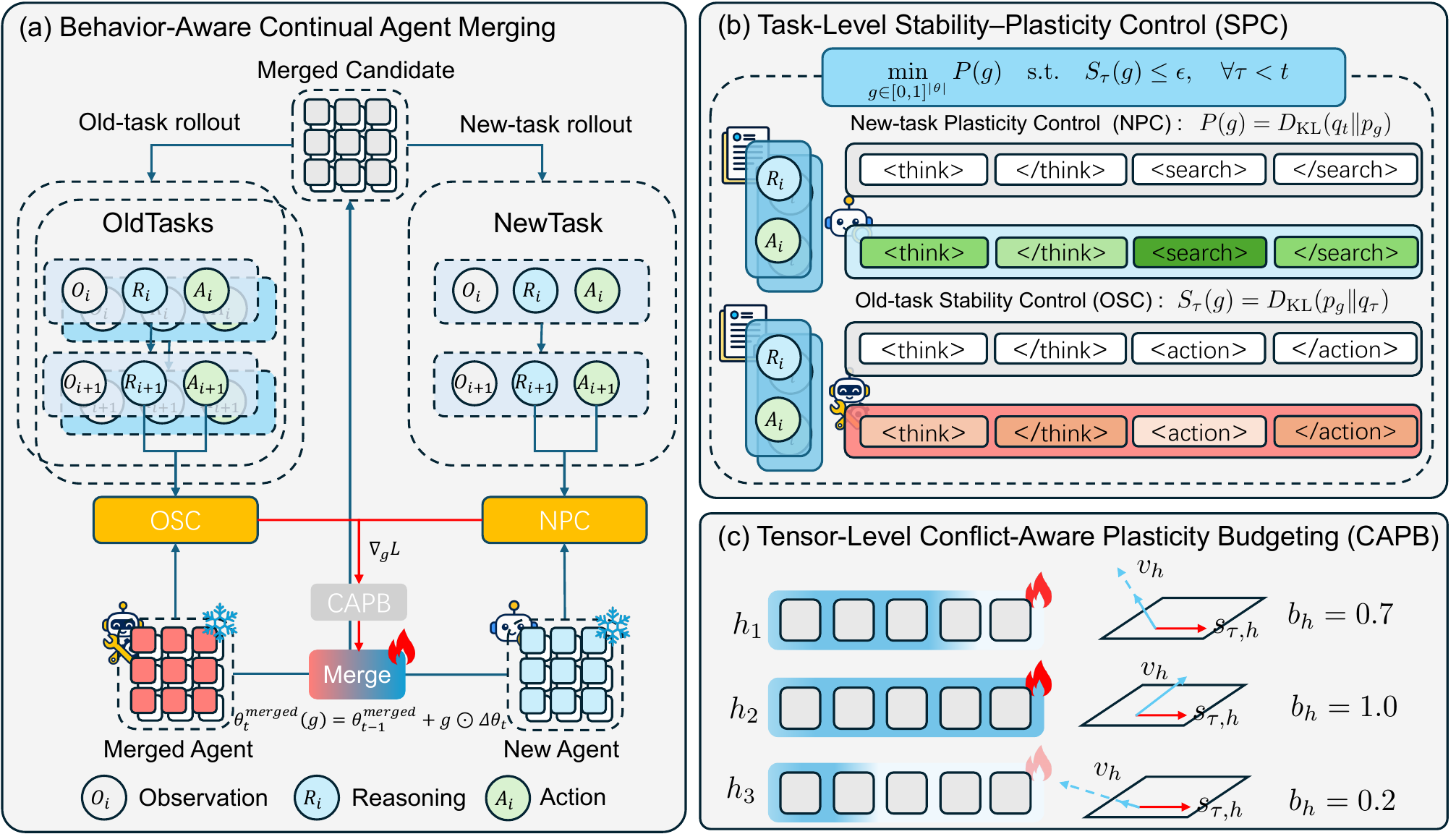}
  \caption{Overview of BACAM. (a) BACAM constructs a merged candidate from the current merged model and incoming expert, collects interaction trajectories, and optimizes the gates through task- and tensor-level control. (b) Task-level stability--plasticity control (SPC) combines forward-KL supervision from the incoming expert with reverse-KL constraints from the current merged model on candidate-generated histories. (c) Tensor-level conflict-aware plasticity budgeting (CAPB) uses local gradient probes to identify conflicts and restrict gate updates toward the incoming expert in conflicting tensors.}
  \label{fig:overview}
\end{figure}

BACAM constructs a merged candidate from $M_{t-1}$ and $E_t$ and rolls it out on the new and earlier tasks (Figure~\ref{fig:overview}). Within Task-Level Stability--Plasticity Control (SPC), \textbf{New-task Plasticity Control (NPC)} uses $E_t$ to guide capability acquisition, while \textbf{Old-task Stability Control (OSC)} uses $M_{t-1}$ to limit behavioral drift on earlier tasks. \textbf{Tensor-Level Conflict-Aware Plasticity Budgeting (CAPB)} restricts gate updates in tensors where these two goals conflict.

To define the merged candidate, let $\theta_{t-1}^{\mathrm{merged}}$ and $\theta_t$ denote the parameters of $M_{t-1}$ and $E_t$, respectively, and set $\Delta\theta_t=\theta_t-\theta_{t-1}^{\mathrm{merged}}$. A parameter-wise gate $g$ determines the candidate's weights:
\begin{equation}
\theta_t^{\mathrm{merged}}(g)
=\theta_{t-1}^{\mathrm{merged}}+g\odot\Delta\theta_t,
\qquad g\in[0,1]^{|\theta|}.
\label{eq:gated-merge}
\end{equation}
The current merged model and incoming expert remain frozen, and only $g$ is optimized. For parameter $j$, $g_j=0$ retains the old-model value and $g_j=1$ selects the new-expert value. Compared with coarse-grained merging coefficients, the parameter-wise gate can express different stability--plasticity trade-offs within the same tensor~\citep{yang2024adamerging,zhang2026expert}.

\subsection{Task-Level Stability--Plasticity Control}
\label{sec:task-level-control}

\paragraph{Behavioral supervision on candidate trajectories.}
The candidate's actions affect subsequent observations, so fixed prompts or expert-generated trajectories may not cover the histories it visits. For each task $\tau\leq t$, we roll out the candidate policy $\pi_{\theta_t^{\mathrm{merged}}(g)}$ in the task environment using instances from $\mathcal{D}_\tau$, obtaining trajectories $\mathcal{R}_\tau$. We write the decision points of each trajectory $\xi$ as $((x_i,y_i))_{i=1}^{|\xi|}$, where $x_i$ contains the interaction history and current observation, and $y_i$ is the candidate's reasoning and action. The incoming expert supervises task $t$; the current merged model supervises tasks $\tau<t$. At response token $k$ of decision point $i$, the candidate and teacher distributions are
\begin{equation}
\begin{aligned}
p_{g,i,k}
&=\pi_{\theta_t^{\mathrm{merged}}(g)}(\cdot\mid x_i,y_{i,<k}),\\
q_{t,i,k}
&=\pi_{\theta_t}(\cdot\mid x_i,y_{i,<k}),\\
q_{\tau,i,k}
&=\pi_{\theta_{t-1}^{\mathrm{merged}}}(\cdot\mid x_i,y_{i,<k}),
\quad \tau<t.
\end{aligned}
\label{eq:student-teacher-distributions}
\end{equation}
At each token, the candidate and teacher are evaluated on the same candidate-generated prefix; only the candidate interacts with the environment~\citep{agarwal2024policy,song2026survey}.

\paragraph{Behavior-aware loss aggregation.}
To form task-level losses from these distributions, we account for the different roles of response tokens. Reasoning text affects the environment indirectly, whereas search queries, function calls, and environment actions directly determine the next observation. Using task-format markers, we partition the response positions at decision point $i$ into critical behavior tokens $C_i$ and remaining tokens $N_i$. Let $\mathcal{B}_i\subseteq\{C_i,N_i\}$ contain the nonempty groups. For a token-level divergence $D$, we average uniformly within each group and then across the available groups:
\begin{equation}
\ell_i^{D}(g;q_\tau)
=\mathbb{E}_{B\in\mathcal{B}_i}
\mathbb{E}_{k\in B}
\left[D\!\left(p_{g,i,k},q_{\tau,i,k}\right)\right].
\label{eq:behavior-aware-token-loss}
\end{equation}
This prevents ordinary text tokens from overwhelming sparse but consequential action tokens. We also give each trajectory equal weight:
\begin{equation}
Z_\tau^{D}(g;q_\tau)
=\mathbb{E}_{\xi\in\mathcal{R}_\tau}
\mathbb{E}_{i=1}^{|\xi|}
\left[\ell_i^{D}(g;q_\tau)\right],
\label{eq:trajectory-equal-loss}
\end{equation}
Within a trajectory, we average over its decision points, so longer interactions do not dominate the loss. We use this aggregation for both $P(g)$ and $S_\tau(g)$, with different KL directions for the new and earlier tasks.

\paragraph{Plasticity and stability losses.}
For NPC, we use forward KL from the incoming expert to the candidate:
\begin{equation}
P(g)=Z_t^{D_{\mathrm{F}}}(g;q_t),
\qquad
D_{\mathrm{F}}(p_{g,i,k},q_{t,i,k})
=D_{\mathrm{KL}}(q_{t,i,k}\|p_{g,i,k}),
\label{eq:plasticity-objective}
\end{equation}
This penalizes the candidate when it assigns insufficient probability to tokens preferred by the incoming expert, encouraging it to acquire the expert's behavior on the histories it visits~\citep{agarwal2024policy}.

For OSC on each earlier task $\tau<t$, we use reverse KL from the candidate to the current merged model:
\begin{equation}
S_\tau(g)
=Z_\tau^{D_{\mathrm{R}}}(g;q_\tau),
\qquad
D_{\mathrm{R}}(p_{g,i,k},q_{\tau,i,k})
=D_{\mathrm{KL}}(p_{g,i,k}\|q_{\tau,i,k}),
\label{eq:stability-objective}
\end{equation}
This penalizes probability mass assigned to tokens that the current merged model considers unlikely. Limiting $S_\tau(g)$ therefore constrains behavioral drift on task $\tau$ and provides the stability signal for preserving its previously integrated capability~\citep{gu2024minillm}.

\paragraph{Task-level objective and optimization.}
To acquire the incoming capability while preserving earlier ones, BACAM minimizes $P(g)$ subject to a behavioral-drift budget $\epsilon$ for each old task:
\begin{equation}
\min_{g\in[0,1]^{|\theta|}} P(g)
\quad\text{s.t.}\quad
S_\tau(g)\leq\epsilon,
\quad \forall \tau<t,
\label{eq:constrained-objective}
\end{equation}
We optimize this constrained objective using a Lagrangian with one nonnegative multiplier $\lambda_\tau$ for each old task:
\begin{equation}
\mathcal{L}(g,\lambda)
=P(g)+\sum_{\tau<t}
\lambda_\tau\bigl(S_\tau(g)-\epsilon\bigr),
\label{eq:lagrangian}
\end{equation}
Gate updates descend $\mathcal{L}$, while projected dual-ascent steps increase $\lambda_\tau$ when $S_\tau(g)>\epsilon$ and decrease it when the drift is below budget. Each old task therefore adjusts its own stability weight as merging proceeds. The update rules are summarized in Algorithm~\ref{alg:bacam} and detailed in Appendix~\ref{sec:optimization-details}.

\subsection{Tensor-Level Conflict-Aware Plasticity Budgeting}
\label{sec:tensor-level-control}

Each task-level multiplier sets one stability weight for an old task across the whole model. It cannot distinguish tensors in which acquiring the new task would disrupt that task. CAPB therefore assigns a separate plasticity budget to each tensor to limit such updates.

\paragraph{Tensor-level conflict estimation.}
We estimate local conflict after a temporary step toward the new task. During each refresh round, the collected trajectories remain fixed while we compute gradients. Let $g_{\mathrm{start}}$ be the current gate and $v=\nabla_gP(g_{\mathrm{start}})$ the new-task gradient. The probe gate is
\begin{equation}
g_{\mathrm{probe}}
=\operatorname{clip}\!\left(
g_{\mathrm{start}}
-\eta_{\mathrm{probe}}
\frac{v}{\|v\|_2}
,0,1\right).
\label{eq:probe-step}
\end{equation}
Here, $\operatorname{clip}(\cdot,0,1)$ clips each gate component to $[0,1]$. Let $v_h$ be the part of $v$ corresponding to tensor $h$. At $g_{\mathrm{probe}}$, we compute the old-task stability gradient $s_{\tau,h}=\nabla_{g_h}S_\tau(g_{\mathrm{probe}})$. A negative inner product $\langle v_h,s_{\tau,h}\rangle$ means that further movement along $-v_h$ would locally increase old-task drift. We measure this conflict by
\begin{equation}
c_{\tau,h}
=\max\!\left(
0,
\frac{-\langle v_h,s_{\tau,h}\rangle}
{\|v_h\|_2\|s_{\tau,h}\|_2+\delta}
\right).
\label{eq:gradient-conflict}
\end{equation}
Here, $\delta>0$ is a small constant for numerical stability.

\paragraph{Plasticity budget construction.}
The conflict score captures direction but not how strongly the old task responds to changes in tensor $h$. We therefore combine $c_{\tau,h}$ with the gradient magnitudes $V_h=\|v_h\|_2$ and $U_{\tau,h}=\|s_{\tau,h}\|_2$:
\begin{equation}
b_{\tau,h}
=1-\frac{c_{\tau,h}U_{\tau,h}}
{V_h+U_{\tau,h}+\delta},
\qquad
b_h=\min_{\tau<t}b_{\tau,h}.
\label{eq:plasticity-budget}
\end{equation}
The budget satisfies $0\leq b_h\leq1$. Tensors with no detected conflict have $b_h=1$; an opposing gradient and stronger old-task sensitivity reduce the budget. Taking the minimum gives each tensor the most restrictive budget across old tasks. The probe only estimates these budgets; the actual gate updates start from $g_{\mathrm{start}}$.

\paragraph{Budgeted gate updates.}
Let $\Delta g$ be the normalized descent step computed from the task-level objective $\mathcal{L}$, as in Algorithm~\ref{alg:bacam}. We apply the tensor budgets only to positive gate updates, which move parameters toward the new expert:
\begin{equation}
\Delta g_{\mathrm{safe},h}
=b_h[\Delta g_h]_+ + [\Delta g_h]_-,
\qquad
g_h\leftarrow\operatorname{clip}\!\left(g_h+\Delta g_{\mathrm{safe},h},0,1\right),
\label{eq:safe-gate-update}
\end{equation}
where $[z]_+=\max(z,0)$ and $[z]_- =\min(z,0)$ are applied elementwise. Negative updates move the gate back toward the old model and are therefore not restricted by the plasticity budget.

\IncMargin{1em}
\begin{algorithm}[t]
\caption{Behavior-Aware Continual Agent Merging}
\label{alg:bacam}
\KwIn{$\theta_{t-1}^{\mathrm{merged}}$, $\theta_t$, unlabeled data $\{\mathcal{D}_\tau\}_{\tau=1}^{t}$, rounds $N$}
Initialize $\Delta\theta_t\gets\theta_t-\theta_{t-1}^{\mathrm{merged}}$, $g\gets\mathbf{0}$, $\lambda_\tau\gets0$ for $\tau<t$\;
\For{$n=1,\ldots,N$}{
    $\theta_t^{\mathrm{merged}}(g)\gets\theta_{t-1}^{\mathrm{merged}}+g\odot\Delta\theta_t$\;
    Collect $\mathcal{R}_\tau$ from $\mathcal{D}_\tau$, $\tau=1,\ldots,t$\;
    Construct tensor-wise plasticity budgets $\{b_h\}$ (Eqs.~\ref{eq:probe-step}--\ref{eq:plasticity-budget})\;
    \For{each gate-update step}{
        $P(g)\gets Z_t^{D_{\mathrm{F}}}(g;q_t)$\;
        $S_\tau(g)\gets Z_\tau^{D_{\mathrm{R}}}(g;q_\tau)$, $\forall\tau<t$\;
        $\mathcal{L}(g,\lambda)\gets P(g)+\sum_{\tau<t}\lambda_\tau(S_\tau(g)-\epsilon)$\;
        $\Delta g\gets-\eta_g\nabla_g\mathcal{L}/\|\nabla_g\mathcal{L}\|_2$\;
        $g_h\gets\operatorname{clip}\!\left(g_h+b_h[\Delta g_h]_+ + [\Delta g_h]_-,0,1\right)$, $\forall h$\;
        $\lambda_\tau\gets\operatorname{clip}\!\left(\lambda_\tau+\eta_\lambda[S_\tau(g)-\epsilon],0,\lambda_{\max}\right)$, $\forall\tau<t$\;
    }
}
\KwRet{$\theta_t^{\mathrm{merged}}(g)$}\;
\end{algorithm}
\DecMargin{1em}

Algorithm~\ref{alg:bacam} summarizes the rollout, budget, and gate-update steps. After $N$ refresh rounds, the learned gate is folded into the model parameters, introducing no additional inference-time parameters or computation.

\section{Experiments}
\label{sec:experiments}

\subsection{Experimental Setup}
\label{sec:experimental-setup}

\paragraph{Backbones.}
We evaluate BACAM with two backbone families, Qwen2.5-7B-Instruct~\citep{qwen2.5} and Llama-3.1-8B-Instruct~\citep{grattafiori2024llama}. The Qwen experts are ReSearch for information retrieval~\citep{chen2025research}, ToolRL for tool use~\citep{qian2026toolrl}, and GiGPO experts for WebShop and ALFWorld~\citep{feng2026group}. The Llama experts are WebShop and ALFWorld GRPO checkpoints from the verl-agent ecosystem~\citep{feng2026group}, xLAM for tool use~\citep{prabhakar2026apigen}, and SSRL for information retrieval~\citep{fan2025ssrl}. Within each family, the four experts share an architecture and tokenizer. The default merging order is WebShop $\rightarrow$ Tool $\rightarrow$ Search $\rightarrow$ ALFWorld (WTSA), with the WebShop expert as the initial model.

\paragraph{Evaluations.}
We evaluate shopping with WebShop~\citep{yao2022webshop}, tool use with BFCL v3 \texttt{multi\_turn\_base}~\citep{patil2025berkeley}, information retrieval with MuSiQue~\citep{trivedi2022musique}, and embodied interaction with ALFWorld~\citep{shridhar2020alfworld}. The primary metric is success rate (SR, \% $\uparrow$) under each benchmark's criterion; for ALFWorld, we report both IID (\texttt{valid\_seen}) and OOD (\texttt{valid\_unseen}) SR. Avg SR is the mean of the five reported columns, counting the two ALFWorld splits separately. We also report WebShop task score (TS $\uparrow$) and average interaction steps (Round $\downarrow$) for WebShop and ALFWorld. All methods use the same held-out instances and benchmark settings. Dataset sizes, scoring rules, and generation settings are given in Appendix~\ref{sec:additional-experimental-details}.

\paragraph{Baselines.}
We compare methods with different access to task instances during merging. Weight averaging~\citep{wortsman2022model}, Task Arithmetic~\citep{ilharco2022editing}, TIES-Merging~\citep{yadav2023ties}, DARE variants~\citep{yu2024language}, TSV-Merge~\citep{gargiulo2025task}, Iso-C~\citep{marczak2025no}, WUDI-Merging~\citep{cheng2025whoever}, RAM and RAM++~\citep{yuan2026behavior}, OPCM~\citep{tang2025merging}, and NUFILT~\citep{qiu2026null} operate on model parameters without task instances. AdaMerging and AdaMerging++~\citep{yang2024adamerging} instead minimize prediction entropy on unlabeled prompts, while Expert Merging~\citep{zhang2026expert} aligns hidden states and output distributions using unlabeled calibration data. We also evaluate the base model and each individual expert. The final-model comparison includes both batch and continual methods using the same four experts. Table~\ref{tab:baseline-protocol} details their merge-time data and supervision.

\paragraph{Implementation details.}
We use 20 candidate-generated episodes per task stream in each round. The gate learning rate is $2\times10^{-2}$ with cosine decay, and the stability budget is $\epsilon=5\times10^{-3}$. The dual update uses $\eta_\lambda=\lambda_{\max}=1$. Training uses $8\times$ RTX 5090 GPUs. Additional settings are provided in Appendix~\ref{sec:additional-experimental-details}.

\subsection{Performance Comparison}
\label{sec:main-results}

\begin{table}[t]
\centering
\caption{Main results on Qwen2.5-7B-Instruct. Scores are reported as SR (\%).}
\label{tab:main-results}
\footnotesize
\renewcommand{\arraystretch}{0.90}
\begin{tabular}{@{}lrrrrrr@{}}
\toprule
Method & WebShop & Tool & Search & \multicolumn{2}{c}{ALFWorld} & Avg $\uparrow$ \\
\cmidrule(lr){5-6}
 & & & & IID & OOD & \\
\midrule
\rowcolor{black!6}
\multicolumn{7}{l}{\textit{Base model and individual experts}} \\
Qwen2.5-7B-Instruct & 0.00 & 7.00 & 12.49 & 20.71 & 15.67 & 11.18 \\
WebShop expert & \cellcolor{yellow!20}78.00 & 4.00 & 7.34 & 24.29 & 20.90 & 26.90 \\
Tool expert & 0.00 & \cellcolor{yellow!20}30.00 & 12.74 & 10.71 & 13.43 & 13.38 \\
Search expert & 0.80 & 11.00 & \cellcolor{yellow!20}24.34 & 9.29 & 10.45 & 11.18 \\
ALFWorld expert & 1.80 & 16.00 & 14.92 & \cellcolor{yellow!20}98.57 & \cellcolor{yellow!20}91.04 & 44.47 \\
\midrule
\rowcolor{black!6}
\multicolumn{7}{l}{\textit{General model merging}} \\
Weight Average & 4.80 & 0.00 & 21.76 & 35.00 & 37.31 & 19.78 \\
Task Arithmetic & 32.00 & \underline{2.00} & 20.82 & 66.43 & 58.21 & 35.89 \\
TIES-Merging & 38.40 & 0.00 & 15.17 & 53.57 & 56.72 & 32.77 \\
TA w/ DARE & 34.00 & 0.00 & 20.38 & 69.29 & 55.22 & 35.78 \\
TIES w/ DARE & 36.20 & 1.00 & 16.96 & 59.29 & 55.97 & 33.88 \\
TSV-Merge & \underline{39.40} & 1.00 & 23.10 & 68.57 & 66.42 & 39.70 \\
Iso-C & 24.60 & 1.00 & \underline{23.85} & 59.29 & 55.97 & 32.94 \\
WUDI-Merging & 38.20 & 0.00 & \textbf{25.14} & \underline{72.14} & \underline{70.15} & \underline{41.13} \\
AdaMerging & 16.80 & 1.00 & 15.27 & 62.86 & 59.70 & 31.13 \\
AdaMerging++ & 21.40 & 1.00 & 18.69 & 62.86 & 60.45 & 32.88 \\
Expert Merging & 5.60 & 1.00 & 18.84 & 28.57 & 32.09 & 17.22 \\
\midrule
\rowcolor{black!6}
\multicolumn{7}{l}{\textit{Agent model merging}} \\
RAM & 34.60 & 1.00 & 21.02 & 56.43 & 52.99 & 33.21 \\
RAM++ & 37.20 & 1.00 & 20.33 & 61.43 & 50.00 & 33.99 \\
\midrule
\rowcolor{black!6}
\multicolumn{7}{l}{\textit{Continual model merging}} \\
OPCM & 22.60 & 1.00 & 21.52 & 46.43 & 41.79 & 26.67 \\
NUFILT & 10.00 & 1.00 & 20.53 & 36.43 & 32.84 & 20.16 \\
\midrule
\rowcolor{blue!9}
\textbf{BACAM (ours)} & \textbf{77.20} & \textbf{38.00} & 23.75 & \textbf{88.57} & \textbf{86.57} & \textbf{62.82} \\
\bottomrule
\end{tabular}
\end{table}

\begin{table}[t]
\centering
\caption{Main results on Llama-3.1-8B-Instruct. Scores are reported as SR (\%).}
\label{tab:llama-main-results}
\footnotesize
\renewcommand{\arraystretch}{0.90}
\begin{tabular}{@{}lrrrrrr@{}}
\toprule
Method & WebShop & Tool & Search & \multicolumn{2}{c}{ALFWorld} & Avg $\uparrow$ \\
\cmidrule(lr){5-6}
 & & & & IID & OOD & \\
\midrule
\rowcolor{black!6}
\multicolumn{7}{l}{\textit{Base model and individual experts}} \\
Llama-3.1-8B-Instruct & 0.00 & 6.00 & 10.66 & 24.29 & 20.90 & 12.37 \\
WebShop expert & \cellcolor{yellow!20}56.60 & 0.00 & 9.32 & 5.00 & 3.73 & 14.93 \\
Tool expert & 0.00 & \cellcolor{yellow!20}51.00 & 8.78 & 0.00 & 0.00 & 11.96 \\
Search expert & 0.00 & 4.00 & \cellcolor{yellow!20}22.76 & 17.86 & 13.43 & 11.61 \\
ALFWorld expert & 2.60 & 3.00 & 10.01 & \cellcolor{yellow!20}75.71 & \cellcolor{yellow!20}60.45 & 30.36 \\
\midrule
\rowcolor{black!6}
\multicolumn{7}{l}{\textit{General model merging}} \\
Weight Average & 0.20 & 0.00 & 22.11 & 14.29 & 9.70 & 9.26 \\
Task Arithmetic & 0.00 & 0.00 & 16.76 & 8.57 & 8.96 & 6.86 \\
TIES-Merging & 0.00 & 0.00 & 19.53 & 7.86 & 10.45 & 7.57 \\
TA w/ DARE & 0.00 & 0.00 & 16.46 & 8.57 & 8.96 & 6.80 \\
TIES w/ DARE & 0.20 & 0.00 & 18.59 & 13.57 & 11.94 & 8.86 \\
TSV-Merge & 0.60 & 0.00 & 23.40 & 21.43 & 20.90 & 13.27 \\
Iso-C & \underline{2.00} & \underline{0.00} & \underline{24.89} & \underline{30.00} & \underline{31.34} & \underline{17.65} \\
WUDI-Merging & 0.60 & 0.00 & \textbf{24.94} & 25.00 & 26.12 & 15.33 \\
\midrule
\rowcolor{black!6}
\multicolumn{7}{l}{\textit{Agent model merging}} \\
RAM & 0.60 & 0.00 & 21.76 & 12.86 & 11.94 & 9.43 \\
RAM++ & 0.60 & 0.00 & 21.86 & 12.86 & 11.19 & 9.30 \\
\midrule
\rowcolor{black!6}
\multicolumn{7}{l}{\textit{Continual model merging}} \\
OPCM & 0.40 & 0.00 & 23.05 & 15.00 & 15.67 & 10.83 \\
NUFILT & 0.00 & 0.00 & 23.10 & 19.29 & 18.66 & 12.21 \\
\midrule
\rowcolor{blue!9}
\textbf{BACAM (ours)} & \textbf{62.40} & \textbf{41.00} & 21.67 & \textbf{62.86} & \textbf{59.70} & \textbf{49.52} \\
\bottomrule
\end{tabular}
\end{table}

\paragraph{Overall performance.}
With the Qwen series, BACAM achieves an Avg SR of 62.82\%, exceeding the strongest measured baseline, WUDI-Merging, by 21.69 percentage points (Table~\ref{tab:main-results}). The largest differences are on Tool and WebShop: BACAM reaches 38.00\% Tool SR, compared with 0--2\% for the evaluated merging baselines, and 77.20\% WebShop SR, compared with 39.40\% for TSV-Merge. On Search, BACAM scores 23.75\%, below WUDI-Merging's 25.14\%.

\paragraph{Comparison with specialists.}
BACAM remains close to the WebShop and Search experts, trailing them by 0.80 and 0.59 percentage points, respectively. Tool SR exceeds its expert by 8.00 points, whereas ALFWorld IID/OOD SR remains 10.00/4.47 points below its expert. Additional TS and Round results are reported in Table~\ref{tab:interaction-results}.

\paragraph{Two- and three-expert merging.}
Appendix Table~\ref{tab:prefix-results} reports the first two merging stages in the default order. After adding Tool, BACAM reaches 32.00\% Tool SR while retaining 78.60\% WebShop SR; after adding Search, these scores are 37.00\% and 78.40\%, with 24.10\% Search SR. The other merging methods reach at most 2.00\% Tool SR at either stage. These results show that BACAM acquires the incoming capabilities while preserving the initial WebShop capability through the first three experts.

\paragraph{Different architectures.}
With the Llama series, BACAM achieves an average score of 49.52\%, outperforming the strongest merging baseline, Iso-C, by 31.87 percentage points (Table~\ref{tab:llama-main-results}). On WebShop, BACAM reaches 62.40\% SR, compared with 2.00\% for Iso-C. On Tool, it reaches 41.00\% SR, while all merging baselines score 0.00\%. On Search, BACAM scores 21.67\%, below WUDI-Merging's 24.94\%. On ALFWorld, BACAM reaches 62.86\% IID and 59.70\% OOD SR, compared with 30.00\% and 31.34\% for Iso-C, respectively.

\FloatBarrier

\subsection{Continual Merging Analysis}
\label{sec:continual-results}

\begin{figure}[h]
\centering
\includegraphics[width=\linewidth]{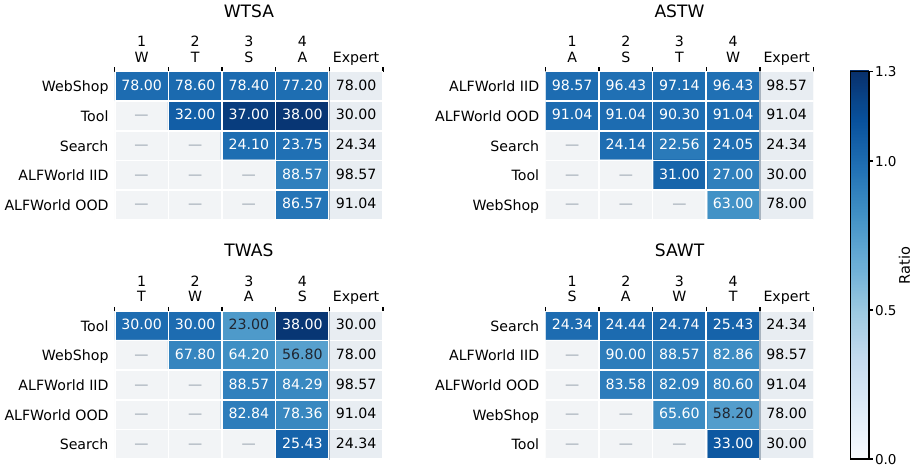}
\caption{BACAM's stage-wise SR (\%) under four arrival orders. W, T, S, and A denote WebShop, Tool, Search, and ALFWorld. Cell values are SR; shading shows SR relative to the corresponding expert. --- marks tasks not yet integrated.}
\label{fig:merging-stages}
\end{figure}

We evaluate BACAM under four expert orders: WTSA, ASTW, TWAS, and SAWT, and track SR after each integration (Figure~\ref{fig:merging-stages}). Appendix Table~\ref{tab:stage-results} gives the full scores.

\paragraph{Acquisition and retention.}
In WTSA, adding Tool raises its SR from the initial WebShop expert's 4.00\% to 32.00\%, while WebShop SR changes from 78.00\% to 78.60\%. Subsequent integrations bring Tool SR to 38.00\%, while final WebShop SR is 77.20\%, only 0.80 points below its initial value. Search changes from 24.10\% at integration to 23.75\% after ALFWorld is added. ALFWorld enters last and reaches 88.57/86.57\% IID/OOD SR. BACAM does not fully acquire the ALFWorld expert's capability at this integration stage.

\paragraph{Effect of merging order.}
Merging order has a substantial but task-dependent effect. Final Avg SR ranges from 56.02\% in SAWT to 62.82\% in WTSA, a difference of 6.80 points. Across the four orders, WebShop varies by 20.40 points, whereas Search varies by only 1.68 points. The initial expert's own-task capability is largely preserved: WebShop in WTSA ends 0.80 points below its expert, and ALFWorld in ASTW ends 2.14/0.00 points below its IID/OOD expert scores. By contrast, adding WebShop later leaves a 10.20--15.00-point gap at integration, and adding ALFWorld later leaves an 8.57--10.00-point IID gap; Tool and Search reach or approach their expert scores. Thus, order mainly affects the acquisition and subsequent retention of certain incoming capabilities rather than uniformly degrading all tasks.

\paragraph{Parameter adaptation.}
Under TWAS, gate movement decreases across successive integrations (0.035, 0.025, and 0.013), indicating increasingly conservative updates along this sequence. MLP tensors account for 72--77\% of total movement, while value projections have the largest average gate changes within attention. LayerNorm receives the tightest plasticity budgets. During ALFWorld integration, BACAM further protects WebShop more strongly than Tool in middle-layer value projections (Appendix Figure~\ref{fig:gate-value-conflict}). These patterns show that adaptation depends on both stage and tensor type; further diagnostics appear in Appendix~\ref{sec:gate-diagnostics}.


\subsection{Ablation on Each Component}
\label{sec:ablations}

\begin{table}[t]
\centering
\caption{Four-expert ablation SR (\%) under WTSA.}
\label{tab:ablations}
\footnotesize
\renewcommand{\arraystretch}{0.90}
\begin{tabular}{@{}lrrrrrr@{}}
\toprule
Method & WebShop & Tool & Search & \multicolumn{2}{c}{ALFWorld} & Avg $\uparrow$ \\
\cmidrule(lr){5-6}
 & & & & IID & OOD & \\
\midrule
\rowcolor{black!6}
\multicolumn{7}{l}{\textit{(1) Task-level behavioral constraints}} \\
Equal task weighting & 77.80 & 34.00 & 23.95 & 90.00 & 81.34 & 61.42 \\
Forward KL for all tasks & 76.80 & 30.00 & 23.30 & 89.29 & 85.07 & 60.89 \\
Reverse KL for all tasks & 77.00 & 36.00 & 24.05 & 87.86 & 85.82 & 62.14 \\
Uniform token weighting & 77.00 & 35.00 & 23.55 & 87.14 & 84.33 & 61.40 \\
\midrule
\rowcolor{black!6}
\multicolumn{7}{l}{\textit{(2) Tensor-level plasticity budgets}} \\
w/o CAPB & 77.40 & 34.00 & 22.81 & 87.86 & 85.82 & 61.58 \\
\midrule
\rowcolor{blue!9}
\textbf{BACAM (ours)} & 77.20 & 38.00 & 23.75 & 88.57 & 86.57 & 62.82 \\
\bottomrule
\end{tabular}
\end{table}

We assess BACAM's two levels of control after the full WebShop--Tool--Search--ALFWorld sequence (Table~\ref{tab:ablations}). The variants retain behavioral supervision while changing task-level loss weights, KL directions, token weighting, or tensor-level budgets throughout merging.

(1) \textbf{Task-level behavioral constraints.} Replacing task-level stability--plasticity control with equal weighting of new- and old-task losses yields 61.42\% Avg SR. Using forward KL for all tasks slightly improves ALFWorld IID SR from 88.57\% to 89.29\%, but lowers its OOD SR from 86.57\% to 85.07\% and reduces Tool SR from 38.00\% to 30.00\%. Reverse KL for all tasks slightly improves Search SR but lowers SR on the other tasks. Neither uniform KL direction matches BACAM's overall performance: BACAM achieves 62.82\% Avg SR, compared with 60.89\% for forward KL and 62.14\% for reverse KL. Replacing token-group weighting with uniform token weighting reduces Avg SR to 61.40\%, with the largest decreases on Tool and ALFWorld OOD. These comparisons support using different KL directions for acquisition and preservation, together with explicit weighting of behavior-critical tokens.

(2) \textbf{Tensor-level plasticity budgets.} Removing the conflict budget while retaining task-level constraints reduces Avg SR from 62.82\% to 61.58\%. The largest decrease is on Tool, from 38.00\% to 34.00\%; Search and ALFWorld SR also decline, while WebShop SR rises slightly from 77.20\% to 77.40\%. The overall improvement with the budget suggests that tensor-level conflict control complements the task-level objective, although its benefit is not uniform across tasks.

\begin{table}[htbp]
\centering
\caption{Ablation of data amount.}
\label{tab:data-amount-ablation}
\footnotesize
\setlength{\tabcolsep}{3pt}
\renewcommand{\arraystretch}{1.1}
\begin{tabular}{@{}llrrrrrr@{}}
\toprule
Method & Data per task & WebShop & Tool & Search & \multicolumn{2}{c}{ALFWorld} & Avg \\
\cmidrule(lr){6-7}
 & & & & & IID & OOD & \\
\midrule
BACAM (ours) & 1 episode & 77.80 & 14.00 & 20.72 & 52.86 & 54.48 & 43.97 \\
BACAM (ours) & 6 episodes & 76.80 & 28.00 & 23.15 & 86.43 & 79.10 & 58.70 \\
\rowcolor{blue!9}
BACAM (ours) & 20 episodes & 77.20 & 38.00 & 23.75 & 88.57 & 86.57 & 62.82 \\
\bottomrule
\end{tabular}
\end{table}

(3) \textbf{Amount of behavioral supervision.} Table~\ref{tab:data-amount-ablation} compares BACAM with 1, 6, and 20 episodes per task. Avg SR increases from 43.97\% to 58.70\% and 62.82\%, respectively. These results show that broader coverage of behavioral supervision improves merging performance.

\begin{wraptable}{r}{0.48\linewidth}
\centering
\caption{Avg SR (\%) across stability budgets and CAPB settings.}
\label{tab:stability-budget-ablation}
\footnotesize
\setlength{\tabcolsep}{3pt}
\begin{tabular}{@{}lrrrr@{}}
\toprule
CAPB & $\epsilon=0.001$ & $0.005$ & $0.01$ & $0.05$ \\
\midrule
on & 62.11 & \textbf{62.82} & 61.00 & 58.76 \\
off & - & 61.58 & - & - \\
\bottomrule
\end{tabular}
\end{wraptable}

(4) \textbf{Stability budget.} Table~\ref{tab:stability-budget-ablation} evaluates the stability budget $\epsilon$ with CAPB enabled. The default setting, $\epsilon=0.005$, achieves the highest Avg SR among the tested values. At this setting, removing CAPB reduces Avg SR from 62.82\% to 61.58\%, supporting the complementary roles of task-level stability control and tensor-level conflict control.

\WFclear
\section{Conclusion}
\label{sec:conclusion}

We studied continual agent merging in multi-turn interactive environments. Our method, BACAM, learns parameter-wise merging gates through task-level behavioral constraints and tensor-level conflict-aware plasticity budgets. Supervision on candidate-generated trajectories guides capability acquisition and preservation, while the budgets restrict updates in conflicting tensors. Experiments across four interactive tasks show that BACAM outperforms the evaluated merging baselines across different expert counts. In the default four-expert order, BACAM achieves 62.82\% average success rate, exceeding the strongest baseline by 21.69 percentage points. These findings support the promise of behavior-guided merging for interactive agents, while identifying robustness to expert merging order as a key direction for future work.

\subsection*{AI use statement}

Generative AI tools assisted with improving the clarity and readability of the manuscript. Their suggestions helped refine wording, simplify complex sentences, and improve transitions between paragraphs. The tools also assisted in identifying potentially relevant literature for the background and related-work discussion. The authors selected the final wording and references and take responsibility for the content of the paper.

\subsection*{Reproducibility statement}

Section~\ref{sec:method} specifies the merging objective, update rules, and algorithm. Section~\ref{sec:experimental-setup} describes the experts, benchmarks, metrics, and implementation settings. Appendix~\ref{sec:additional-experimental-details} gives the evaluation protocol and merging configurations; Appendix~\ref{sec:additional-results} reports additional results; and Appendix~\ref{sec:additional-analysis} provides gate and conflict diagnostics.

\bibliography{iclr2027_conference}

@article{shao2024deepseekmath,
  title={Deepseekmath: Pushing the limits of mathematical reasoning in open language models},
  author={Shao, Zhihong and Wang, Peiyi and Zhu, Qihao and Xu, Runxin and Song, Junxiao and Bi, Xiao and Zhang, Haowei and Zhang, Mingchuan and Li, YK and Wu, Yang and others},
  journal={arXiv preprint arXiv:2402.03300},
  year={2024}
}

@article{yao2022react,
  title={{ReAct}: Synergizing reasoning and acting in language models},
  author={Yao, Shunyu and Zhao, Jeffrey and Yu, Dian and Du, Nan and Shafran, Izhak and Narasimhan, Karthik and Cao, Yuan},
  journal={arXiv preprint arXiv:2210.03629},
  year={2022}
}

@inproceedings{liu2024agentbench,
  title={{AgentBench}: Evaluating {LLMs} as agents},
  author={Liu, Xiao and Yu, Hao and Zhang, Hanchen and Xu, Yifan and Lei, Xuanyu and Lai, Hanyu and Gu, Yu and Ding, Hangliang and Men, Kaiwen and Yang, Kejuan and others},
  booktitle={International Conference on Learning Representations},
  volume={2024},
  pages={52989--53046},
  year={2024}
}

@inproceedings{zeng2024agenttuning,
  title={{AgentTuning}: Enabling generalized agent abilities for {LLMs}},
  author={Zeng, Aohan and Liu, Mingdao and Lu, Rui and Wang, Bowen and Liu, Xiao and Dong, Yuxiao and Tang, Jie},
  booktitle={Findings of the Association for Computational Linguistics: ACL 2024},
  pages={3053--3077},
  year={2024}
}

@inproceedings{chen2026scaling,
  title={Scaling agent learning via experience synthesis},
  author={Chen, Zhaorun and Zhao, Zhuokai and Zhang, Kai and Liu, Bo and Qi, Qi and Wu, Yifan and Kalluri, Tarun and Cao, Xuefei and Xiong, Yuanhao and Tong, Haibo and others},
  booktitle={International Conference on Learning Representations},
  volume={2026},
  pages={121394--121420},
  year={2026}
}

@article{luo2025agent,
  title={Agent lightning: Train any ai agents with reinforcement learning},
  author={Luo, Xufang and Zhang, Yuge and He, Zhiyuan and Wang, Zilong and Zhao, Siyun and Li, Dongsheng and Qiu, Luna K and Yang, Yuqing},
  journal={arXiv preprint arXiv:2508.03680},
  year={2025}
}

@article{liu2025gem,
  title={Gem: A gym for agentic llms},
  author={Liu, Zichen and Sims, Anya and Duan, Keyu and Chen, Changyu and Yu, Simon and Zhou, Xiangxin and Xu, Haotian and Xiong, Shaopan and Liu, Bo and Tan, Chenmien and others},
  journal={arXiv preprint arXiv:2510.01051},
  year={2025}
}

@article{yang2026model,
  title={Model merging in llms, mllms, and beyond: Methods, theories, applications, and opportunities},
  author={Yang, Enneng and Shen, Li and Guo, Guibing and Wang, Xingwei and Cao, Xiaochun and Zhang, Jie and Tao, Dacheng},
  journal={ACM Computing Surveys},
  volume={58},
  number={8},
  pages={1--41},
  year={2026},
  publisher={ACM New York, NY}
}

@article{yao2022webshop,
  title={{WebShop}: Towards scalable real-world web interaction with grounded language agents},
  author={Yao, Shunyu and Chen, Howard and Yang, John and Narasimhan, Karthik},
  journal={Advances in Neural Information Processing Systems},
  volume={35},
  pages={20744--20757},
  year={2022}
}

@article{shridhar2020alfworld,
  title={{ALFWorld}: Aligning text and embodied environments for interactive learning},
  author={Shridhar, Mohit and Yuan, Xingdi and C{\^o}t{\'e}, Marc-Alexandre and Bisk, Yonatan and Trischler, Adam and Hausknecht, Matthew},
  journal={arXiv preprint arXiv:2010.03768},
  year={2020}
}

@article{chen2025research,
  title={{ReSearch}: Learning to reason with search for {LLMs} via reinforcement learning},
  author={Chen, Mingyang and Li, Tianpeng and Sun, Haoze and Zhou, Yijie and Zhu, Chenzheng and Wang, Haofen and Pan, Jeff Z and Zhang, Wen and Chen, Huajun and Yang, Fan and others},
  journal={arXiv preprint arXiv:2503.19470},
  year={2025}
}

@article{qian2026toolrl,
  title={{ToolRL}: Reward is all tool learning needs},
  author={Qian, Cheng and Acikgoz, Emre Can and He, Qi and Wang, Hongru and Chen, Xiusi and Hakkani-Tur, Dilek and Tur, Gokhan and Ji, Heng},
  journal={Advances in Neural Information Processing Systems},
  volume={38},
  pages={105523--105553},
  year={2025}
}

@article{feng2026group,
  title={Group-in-Group Policy Optimization for {LLM} Agent Training},
  author={Feng, Lang and Xue, Zhenghai and Liu, Tingcong and An, Bo},
  journal={Advances in Neural Information Processing Systems},
  volume={38},
  pages={46375--46408},
  year={2025}
}

@article{qwen2.5,
    title   = {{Qwen2.5} Technical Report},
    author  = {An Yang and Baosong Yang and Beichen Zhang and Binyuan Hui and Bo Zheng and Bowen Yu and Chengyuan Li and Dayiheng Liu and Fei Huang and Haoran Wei and Huan Lin and Jian Yang and Jianhong Tu and Jianwei Zhang and Jianxin Yang and Jiaxi Yang and Jingren Zhou and Junyang Lin and Kai Dang and Keming Lu and Keqin Bao and Kexin Yang and Le Yu and Mei Li and Mingfeng Xue and Pei Zhang and Qin Zhu and Rui Men and Runji Lin and Tianhao Li and Tingyu Xia and Xingzhang Ren and Xuancheng Ren and Yang Fan and Yang Su and Yichang Zhang and Yu Wan and Yuqiong Liu and Zeyu Cui and Zhenru Zhang and Zihan Qiu},
    journal = {arXiv preprint arXiv:2412.15115},
    year    = {2024}
}

@inproceedings{patil2025berkeley,
  title={The {Berkeley Function Calling Leaderboard (BFCL)}: From tool use to agentic evaluation of large language models},
  author={Patil, Shishir G and Mao, Huanzhi and Yan, Fanjia and Ji, Charlie Cheng-Jie and Suresh, Vishnu and Stoica, Ion and Gonzalez, Joseph E},
  booktitle={Forty-second International Conference on Machine Learning},
  year={2025}
}

@article{trivedi2022musique,
  title={{MuSiQue}: Multihop Questions via Single-hop Question Composition},
  author={Trivedi, Harsh and Balasubramanian, Niranjan and Khot, Tushar and Sabharwal, Ashish},
  journal={Transactions of the Association for Computational Linguistics},
  volume={10},
  pages={539--554},
  year={2022},
  publisher={MIT Press One Broadway, 12th Floor, Cambridge, Massachusetts 02142, USA~…}
}

@inproceedings{wortsman2022model,
  title={Model soups: averaging weights of multiple fine-tuned models improves accuracy without increasing inference time},
  author={Wortsman, Mitchell and Ilharco, Gabriel and Gadre, Samir Ya and Roelofs, Rebecca and Gontijo-Lopes, Raphael and Morcos, Ari S and Namkoong, Hongseok and Farhadi, Ali and Carmon, Yair and Kornblith, Simon and others},
  booktitle={International conference on machine learning},
  pages={23965--23998},
  year={2022},
  organization={Pmlr}
}

@article{ilharco2022editing,
  title={Editing models with task arithmetic},
  author={Ilharco, Gabriel and Ribeiro, Marco Tulio and Wortsman, Mitchell and Gururangan, Suchin and Schmidt, Ludwig and Hajishirzi, Hannaneh and Farhadi, Ali},
  journal={arXiv preprint arXiv:2212.04089},
  year={2022}
}

@article{yadav2023ties,
  title={{TIES-Merging}: Resolving interference when merging models},
  author={Yadav, Prateek and Tam, Derek and Choshen, Leshem and Raffel, Colin A and Bansal, Mohit},
  journal={Advances in neural information processing systems},
  volume={36},
  pages={7093--7115},
  year={2023}
}

@inproceedings{yu2024language,
  title={Language Models are {Super Mario}: Absorbing Abilities from Homologous Models as a Free Lunch},
  author={Yu, Le and Yu, Bowen and Yu, Haiyang and Huang, Fei and Li, Yongbin},
  booktitle={Proceedings of the 41st International Conference on Machine Learning},
  volume={235},
  pages={57755--57775},
  year={2024}
}

@inproceedings{gargiulo2025task,
  title={Task singular vectors: Reducing task interference in model merging},
  author={Gargiulo, Antonio Andrea and Crisostomi, Donato and Bucarelli, Maria Sofia and Scardapane, Simone and Silvestri, Fabrizio and Rodola, Emanuele},
  booktitle={2025 IEEE/CVF Conference on Computer Vision and Pattern Recognition (CVPR)},
  pages={18695--18705},
  year={2025},
  organization={IEEE}
}

@article{marczak2025no,
  title={No task left behind: Isotropic model merging with common and task-specific subspaces},
  author={Marczak, Daniel and Magistri, Simone and Cygert, Sebastian and Twardowski, Bart{\l}omiej and Bagdanov, Andrew D and Van De Weijer, Joost},
  journal={arXiv preprint arXiv:2502.04959},
  year={2025}
}

@article{cheng2025whoever,
  title={Whoever started the interference should end it: Guiding data-free model merging via task vectors},
  author={Cheng, Runxi and Xiong, Feng and Wei, Yongxian and Zhu, Wanyun and Yuan, Chun},
  journal={arXiv preprint arXiv:2503.08099},
  year={2025}
}

@inproceedings{yang2024adamerging,
  title={{AdaMerging}: Adaptive model merging for multi-task learning},
  author={Yang, Enneng and Wang, Zhenyi and Shen, Li and Liu, Shiwei and Guo, Guibing and Wang, Xingwei and Tao, Dacheng},
  booktitle={International Conference on Learning Representations},
  volume={2024},
  pages={22743--22763},
  year={2024}
}

@inproceedings{zhang2026expert,
  title={Expert merging: Model merging with unsupervised expert alignment and importance-guided layer chunking},
  author={Zhang, Dengming and Ma, Xiaowen and Ni, Zhenliang and Wu, Zhenkai and Shu, Han and Jiang, Xin and Chen, Xinghao},
  booktitle={International Conference on Learning Representations},
  volume={2026},
  pages={21780--21803},
  year={2026}
}

@inproceedings{yuan2026behavior,
  title={Behavior knowledge merge in reinforced agentic models},
  author={Yuan, Xiangchi and Shi, Dachuan and Zhang, Chunhui and Liu, Zheyuan and Yao, Shenglong and Vosoughi, Soroush and Lee, Wenke},
  booktitle={Proceedings of the 64th Annual Meeting of the Association for Computational Linguistics (Volume 1: Long Papers)},
  pages={33007--33028},
  year={2026}
}

@article{wang2026m2a,
  title={{M2A}: Synergizing Mathematical and Agentic Reasoning in Large Language Models},
  author={Wang, Junjian and Zhou, Xin and Xu, Qiran and Zhan, Kun},
  journal={arXiv preprint arXiv:2605.09879},
  year={2026}
}

@article{feng2026arm,
  title={{ARM}: Role-Conditioned Neuron Transplantation for Training-Free Generalist {LLM} Agent Merging},
  author={Feng, Zhuoka and Chen, Kang and Zhao, Sihan and Xiong, Kai and Wang, Yaoning and Yu, Minshen and Nian, Junjie and Xiao, Changyi and Cao, Yixin and Jiang, Yugang},
  journal={arXiv preprint arXiv:2601.07309},
  year={2026}
}

@inproceedings{agarwal2024policy,
  title={On-policy distillation of language models: Learning from self-generated mistakes},
  author={Agarwal, Rishabh and Vieillard, Nino and Zhou, Yongchao and Stanczyk, Piotr and Ramos Garea, Sabela and Geist, Matthieu and Bachem, Olivier},
  booktitle={International Conference on Learning Representations},
  volume={2024},
  pages={21246--21263},
  year={2024}
}

@article{song2026survey,
  title={A survey of on-policy distillation for large language models},
  author={Song, Mingyang and Zheng, Mao},
  journal={arXiv preprint arXiv:2604.00626},
  year={2026}
}

@inproceedings{gu2024minillm,
  title={{MiniLLM}: Knowledge distillation of large language models},
  author={Gu, Yuxian and Dong, Li and Wei, Furu and Huang, Minlie},
  booktitle={International Conference on Learning Representations},
  volume={2024},
  pages={32694--32717},
  year={2024}
}

@article{tang2025merging,
  title={Merging models on the fly without retraining: A sequential approach to scalable continual model merging},
  author={Tang, Anke and Yang, Enneng and Shen, Li and Luo, Yong and Hu, Han and Du, Bo and Tao, Dacheng},
  journal={arXiv preprint arXiv:2501.09522},
  year={2025}
}

@article{qiu2025mingle,
  title={MINGLE: Mixture of Null-Space Gated Low-Rank Experts for Test-Time Continual Model Merging},
  author={Qiu, Zihuan and Xu, Yi and He, Chiyuan and Meng, Fanman and Xu, Linfeng and Wu, Qingbo and Li, Hongliang},
  journal={Advances in Neural Information Processing Systems},
  volume={38},
  pages={143841--143878},
  year={2025}
}

@inproceedings{qiu2026null,
  title={Null-space filtering for data-free continual model merging: Preserving stability, promoting plasticity},
  author={Qiu, Zihuan and Wang, Lei and Cao, Yang and Zhang, Runtong and Su, Bing and Xu, Yi and Meng, Fanman and Xu, Linfeng and Wu, Qingbo and Li, Hongliang},
  booktitle={International Conference on Learning Representations},
  volume={2026},
  pages={128492--128518},
  year={2026}
}

@article{yang2025continual,
  title={Continual model merging without data: Dual projections for balancing stability and plasticity},
  author={Yang, Enneng and Tang, Anke and Shen, Li and Guo, Guibing and Wang, Xingwei and Cao, Xiaochun and Zhang, Jie},
  journal={Advances in Neural Information Processing Systems},
  volume={38},
  pages={39275--39305},
  year={2025}
}

@inproceedings{marouf2025ask,
  title={Ask and remember: A questions-only replay strategy for continual visual question answering},
  author={Marouf, Imad Eddine and Tartaglione, Enzo and Lathuili{\`e}re, St{\'e}phane and Van De Weijer, Joost},
  booktitle={2025 IEEE/CVF International Conference on Computer Vision (ICCV)},
  pages={18078--18089},
  year={2025},
  organization={IEEE}
}

@article{barres2025tau2bench,
  title={{\(\tau^2\)}-Bench: Evaluating Conversational Agents in a Dual-Control Environment},
  author={Barres, Victor and Dong, Honghua and Ray, Soham and Si, Xujie and Narasimhan, Karthik},
  journal={arXiv preprint arXiv:2506.07982},
  year={2025}
}

@article{li2025simia,
  title={Simulating Environments with Reasoning Models for Agent Training},
  author={Li, Yuetai and Inan, Huseyin A and Yue, Xiang and Chen, Wei-Ning and Wutschitz, Lukas and Kulkarni, Janardhan and Poovendran, Radha and Sim, Robert and Rajmohan, Saravan},
  journal={arXiv preprint arXiv:2511.01824},
  year={2025}
}

@article{grattafiori2024llama,
  title={The llama 3 herd of models},
  author={Grattafiori, Aaron and Dubey, Abhimanyu and Jauhri, Abhinav and Pandey, Abhinav and Kadian, Abhishek and Al-Dahle, Ahmad and Letman, Aiesha and Mathur, Akhil and Schelten, Alan and Vaughan, Alex and others},
  journal={arXiv preprint arXiv:2407.21783},
  year={2024}
}

@article{prabhakar2026apigen,
  title={Apigen-mt: Agentic pipeline for multi-turn data generation via simulated agent-human interplay},
  author={Prabhakar, Akshara and Liu, Zuxin and Zhu, Ming and Zhang, Jianguo and Awalgaonkar, Tulika Manoj and Wang, Shiyu and Liu, Zhiwei and Chen, Haolin and Hoang, Thai and Niebles, Juan Carlos and others},
  journal={Advances in Neural Information Processing Systems},
  volume={38},
  year={2026}
}

@article{fan2025ssrl,
  title={Ssrl: Self-search reinforcement learning},
  author={Fan, Yuchen and Zhang, Kaiyan and Zhou, Heng and Zuo, Yuxin and Chen, Yanxu and Fu, Yu and Long, Xinwei and Zhu, Xuekai and Jiang, Che and Zhang, Yuchen and others},
  journal={arXiv preprint arXiv:2508.10874},
  year={2025}
}
\bibliographystyle{iclr2027_conference}

\appendix
\section{Formal Setup and Optimization Details}
\label{sec:preliminaries}

\subsection{Continual Merging Setup}
We consider $T$ ordered tasks whose experts share the same architecture and tokenizer. We initialize $M_1=E_1$ and perform merging stages $t=2,\ldots,T$. At stage $t$, task indices $\tau<t$ denote previously integrated tasks and $\tau=t$ denotes the incoming task. Each $\mathcal{D}_\tau$ contains unlabeled instances for task $\tau$. BACAM has access to these data and their interaction environments, the current merged model $M_{t-1}$, and the incoming expert $E_t$. The goal is to construct a unified model $M_t$ that acquires the capability of $E_t$ while retaining the capabilities of $M_{t-1}$:
\begin{equation}
M_t = \operatorname{Merge}(M_{t-1}, E_t).
\label{eq:continual-merging}
\end{equation}
This objective involves two competing requirements. \emph{Plasticity} requires the merged model to adapt its decision rule to the current task, whereas \emph{stability} requires the update to preserve the decision rules that support previously integrated tasks. Because $M_t$ serves as the current model at the next stage, degradation at any stage can accumulate along the merging sequence.

For interactive agents, neither parameter distance nor output discrepancies on isolated histories adequately characterize this objective. An action taken early in an interaction changes the subsequent environment feedback and, consequently, the histories visited later. Continual model merging must therefore constrain the behavioral changes induced by parameter updates, rather than parameter changes alone. We use policy-generated interaction trajectories as the behavioral representation that connects parameter-space optimization to task-level capabilities.

\subsection{Policy-Induced Interaction Trajectories}
Following the ReAct paradigm~\citep{yao2022react}, we represent each interaction step as an observation--reasoning--action triplet, as illustrated in Figure~\ref{fig:overview}. At step $i$, the agent receives an observation $o_i$, produces a reasoning trace $r_i$, and executes an action $a_i$. The environment then returns the next observation:
\begin{equation}
 r_i \sim \pi_\theta(\cdot \mid h_i), \qquad
 a_i \sim \pi_\theta(\cdot \mid h_i, r_i), \qquad
 o_{i+1} \sim \mathcal{E}^{\tau}(\cdot \mid h_i, a_i),
\label{eq:react-step}
\end{equation}
where $h_i=(o_0,r_0,a_0,\ldots,o_i)$ is the interaction history and $\mathcal{E}^{\tau}$ denotes the dynamics of task environment $\tau$. A complete trajectory is therefore
\begin{equation}
\xi=(o_0,r_0,a_0,\ldots,o_{L-1},r_{L-1},a_{L-1},o_L).
\label{eq:agent-trajectory}
\end{equation}
The policy and environment jointly induce the trajectory distribution
\begin{equation}
\mathbb{P}_{\pi_\theta}^{\tau}(\xi)
= \rho_0^{\tau}(o_0)
\prod_{i=0}^{L-1}
\pi_\theta(r_i \mid h_i)
\pi_\theta(a_i \mid h_i,r_i)
\mathcal{E}^{\tau}(o_{i+1} \mid h_i,a_i),
\label{eq:trajectory-distribution}
\end{equation}
where $\rho_0^{\tau}$ is the initial-observation distribution. For an autoregressive language model, the reasoning and action at step $i$ form a token sequence $z_i=(r_i,a_i)=(y_{i,1},\ldots,y_{i,L_i})$, whose probability factorizes as
\begin{equation}
\pi_\theta(r_i,a_i \mid h_i)
= \prod_{\ell=1}^{L_i}
\pi_\theta(y_{i,\ell} \mid h_i, y_{i,<\ell}).
\label{eq:action-probability}
\end{equation}
Agent behavior is thus characterized by the distribution over observation--reasoning--action trajectories, rather than by an isolated model output. An incorrect action changes the next observation and can redirect all subsequent reasoning and actions.

\subsection{Gate and Dual Updates}
\label{sec:optimization-details}
At the beginning of each refresh round, the current candidate generates trajectories for the new and previously integrated tasks. These trajectories and their teacher distributions remain fixed during the gate updates in that round. Gradients are propagated through the candidate parameters but not through trajectory collection. The local conflict probe estimates the tensor budgets without updating the optimizer or dual variables, and the gate is restored before the actual updates.

For completeness, the projected dual update used for each old-task constraint is
\begin{equation}
\lambda_\tau\leftarrow
\operatorname{clip}\!\left(
\lambda_\tau+\eta_\lambda[S_\tau(g)-\epsilon],
0,\lambda_{\max}
\right).
\label{eq:dual-update}
\end{equation}
The parameterization in Equation~\ref{eq:gated-merge} gives the gate gradient by the chain rule:
\begin{equation}
\nabla_g\mathcal{L}
=\Delta\theta_t\odot\nabla_{\theta_t^{\mathrm{merged}}(g)}\mathcal{L}.
\label{eq:gate-gradient}
\end{equation}
The normalized descent step before applying tensor budgets is
\begin{equation}
\Delta g=-\eta_g\frac{\nabla_g\mathcal{L}(g,\lambda)}{\left\|\nabla_g\mathcal{L}(g,\lambda)\right\|_2}.
\label{eq:base-gate-update}
\end{equation}

\section{Experimental Protocol and Merging Configurations}
\label{sec:additional-experimental-details}

\subsection{Model Checkpoints}

\subsubsection{Qwen}

\paragraph{Shared backbone.} We use Qwen2.5-7B-Instruct as the common backbone and tokenizer for the four experts and evaluate the unmerged checkpoint as the base model.\footnote{\href{https://huggingface.co/Qwen/Qwen2.5-7B-Instruct}{\nolinkurl{Qwen/Qwen2.5-7B-Instruct}}}

\paragraph{WebShop expert.} The GiGPO WebShop checkpoint is specialized for interactive shopping and serves as the initial model in the default WTSA order. GiGPO estimates advantages at the episode level and groups actions taken from matching states to estimate step-level advantages~\citep{feng2026group}.\footnote{\href{https://huggingface.co/langfeng01/GiGPO-Qwen2.5-7B-Instruct-WebShop}{\nolinkurl{langfeng01/GiGPO-Qwen2.5-7B-Instruct-WebShop}}}

\paragraph{Tool expert.} The ToolRL checkpoint uses cold-start GRPO to learn tool calling. We evaluate it on the multi-turn subset of BFCL v3. ToolRL uses structured rewards for call format and correctness, including function names and arguments~\citep{qian2026toolrl}.\footnote{\href{https://huggingface.co/emrecanacikgoz/Qwen2.5-7B-Instruct-ToolRL-grpo-cold}{\nolinkurl{emrecanacikgoz/Qwen2.5-7B-Instruct-ToolRL-grpo-cold}}}

\paragraph{Search expert.} The ReSearch checkpoint is specialized for reasoning with retrieval. We evaluate it on MuSiQue. ReSearch uses reinforcement learning to train models to issue queries during reasoning and use retrieved evidence to answer questions~\citep{chen2025research}.\footnote{\href{https://huggingface.co/agentrl/ReSearch-Qwen-7B-Instruct}{\nolinkurl{agentrl/ReSearch-Qwen-7B-Instruct}}}

\paragraph{ALFWorld expert.} The GiGPO ALFWorld checkpoint is specialized for embodied household tasks. We evaluate it on both the seen and unseen splits. GiGPO uses step-level advantages to assign credit to individual actions in long interactions~\citep{feng2026group}.\footnote{\href{https://huggingface.co/langfeng01/GiGPO-Qwen2.5-7B-Instruct-ALFWorld}{\nolinkurl{langfeng01/GiGPO-Qwen2.5-7B-Instruct-ALFWorld}}}

\subsubsection{Llama}

\paragraph{Shared backbone.} We use Llama-3.1-8B-Instruct as the shared backbone and evaluate its original checkpoint as the base model.\footnote{\href{https://huggingface.co/meta-llama/Llama-3.1-8B-Instruct}{\nolinkurl{meta-llama/Llama-3.1-8B-Instruct}}}

\paragraph{WebShop expert.} The WebShop GRPO checkpoint specializes in interactive shopping and initializes the merged model in the default WTSA order. GRPO estimates advantages by comparing rewards within a group of sampled rollouts~\citep{shao2024deepseekmath}.\footnote{\href{https://huggingface.co/ZHLiu627/verl_agent_webshop-new-GRPO-Llama-3.1-8B-Instruct-150step}{\nolinkurl{ZHLiu627/verl_agent_webshop-new-GRPO-Llama-3.1-8B-Instruct-150step}}}

\paragraph{Tool expert.} We use xLAM-2-8B for function calling and evaluate it on the multi-turn subset of BFCL v3. xLAM-2 learns function calling through supervised training on filtered multi-turn agent trajectories~\citep{prabhakar2026apigen}.\footnote{\href{https://huggingface.co/Salesforce/Llama-xLAM-2-8b-fc-r}{\nolinkurl{Salesforce/Llama-xLAM-2-8b-fc-r}}}

\paragraph{Search expert.} We use the SSRL checkpoint for retrieval-augmented reasoning and evaluate it on MuSiQue. SSRL trains models to generate search evidence during reinforcement learning and supports external retrieval at inference time~\citep{fan2025ssrl}.\footnote{\href{https://huggingface.co/TsinghuaC3I/Llama-3.1-8B-Instruct-SSRL}{\nolinkurl{TsinghuaC3I/Llama-3.1-8B-Instruct-SSRL}}}

\paragraph{ALFWorld expert.} The ALFWorld GRPO checkpoint specializes in embodied household tasks. We evaluate it on both the seen and unseen splits. GRPO uses rewards from sampled interactions to improve the policy for choosing actions across multiple steps.\footnote{\href{https://huggingface.co/ZHLiu627/verl_agent_alfworld-GRPO-int-reward_False-Llama-3.1-8B-Instruct-150step}{\nolinkurl{ZHLiu627/verl_agent_alfworld-GRPO-int-reward_False-Llama-3.1-8B-Instruct-150step}}}

\subsection{Evaluation Protocol}

All models are evaluated on the same fixed sets. All baselines use the same evaluation settings as BACAM. All reported baseline and expert scores come from our own runs in the evaluation framework. Before evaluation, we checked expert performance against the official results to verify reproduction accuracy. WebShop uses 500 synthetic shopping goals over a 1,000-product catalog; an episode succeeds when its task score reaches 1.0. BFCL v3 uses 100 held-out \texttt{multi\_turn\_base} tasks and the official checker: every turn must pass its state and response checks. MuSiQue uses 2,017 held-out questions; Search success requires normalized substring matching of the final answer (Sub-EM). ALFWorld uses all 140 \texttt{valid\_seen} games for IID and 134 \texttt{valid\_unseen} games for OOD evaluation. Round is the mean number of interaction steps across all WebShop or ALFWorld episodes, including failures. Table~\ref{tab:protocol-details} gives the generation settings for each benchmark.

\begin{table}[htbp]
\centering
\caption{Evaluation settings. ALFWorld instance counts are shown as IID/OOD; -- indicates that a setting does not apply or has no fixed limit.}
\label{tab:protocol-details}
\small
\begin{tabular}{@{}lrrrr@{}}
\toprule
Setting & WebShop & Tool & Search & ALFWorld \\
\midrule
Evaluation instances & 500 & 100 & 2,017 & 140/134 \\
Temperature & 0.4 & 0.001 & 0 & 0.4 \\
Max. output tokens & 512 & 4,096 & 8,192 & 512 \\
Context length & 8,192 & 32,768 & 8,192 & 8,192 \\
Recent history steps & 2 & -- & -- & 2 \\
Max. environment steps & 15 & -- & -- & 50 \\
Max. tool steps per turn & -- & 20 & -- & -- \\
Retrieved passages & -- & -- & 5 & -- \\
\bottomrule
\end{tabular}
\end{table}
\FloatBarrier

\subsection{Merging Configurations}
\paragraph{BACAM optimization.}
Table~\ref{tab:optimization-settings} reports the settings shared by the four measured orders. These runs differ in expert arrival order. Training instances are unlabeled, and teacher token distributions provide the behavioral targets. The teacher representation retains the top 32 token probabilities and the aggregate probability of the remaining vocabulary, with distillation temperature 1.0.

\begin{table}[htbp]
\centering
\caption{Optimization settings shared by the measured BACAM runs.}
\label{tab:optimization-settings}
\small
\begin{tabular}{@{}ll@{}}
\toprule
Setting & Value \\
\midrule
Rollout-refresh rounds per stage & 3 \\
Gate updates per round & 20 / 30 / 50 \\
Training episodes per task stream per round & 20 \\
States per update per task stream & 16 \\
Gate learning rate / schedule & $2\times10^{-2}$ / cosine \\
Gate range & $[0,1]$ \\
Stability budget $\epsilon$ & $5\times10^{-3}$ \\
Dual update & $\eta_\lambda=\lambda_{\max}=1$ \\
Teacher top-$k$ / temperature & 32 / 1.0 \\
Training hardware & $8\times$ RTX 5090 GPUs \\
\bottomrule
\end{tabular}
\end{table}
\FloatBarrier

\paragraph{Baseline configurations.}
Table~\ref{tab:baseline-protocol} distinguishes task-instance access from the optimization signal in our implementations. RAM and RAM++ use only task-vector statistics, whereas WUDI-Merging and NUFILT optimize objectives derived from model parameters without task instances. AdaMerging and AdaMerging++ use first-turn prompts without multi-turn environment rollouts. Expert Merging and BACAM both use teacher outputs, but on expert-generated and candidate-generated trajectories, respectively. No ground-truth response labels are used for merging, and all calibration instances are separate from the held-out evaluation set.

\begin{table}[htbp]
\centering
\caption{Task-instance access and optimization signals during merging. Instance-dependent methods use unlabeled data; teacher targets are model-generated.}
\label{tab:baseline-protocol}
\small
\begin{tabular}{@{}>{\raggedright\arraybackslash}p{0.25\linewidth}>{\raggedright\arraybackslash}p{0.32\linewidth}>{\raggedright\arraybackslash}p{0.35\linewidth}@{}}
\toprule
Method & Task instances & Optimization signal \\
\midrule
RAM / RAM++ & None & Shared/unique task-vector statistics \\
Other instance-free baselines & None & Parameter-derived rules or objectives; no sample-level targets \\
AdaMerging / AdaMerging++ & Unlabeled first-turn prompts & Prediction entropy; no teacher targets \\
Expert Merging & Expert-generated trajectories (20 episodes/task) & Expert hidden states and output distributions \\
BACAM & Candidate-generated trajectories (20 episodes/task/round) & Token distributions from the incoming expert and current merged model \\
\bottomrule
\end{tabular}
\end{table}

\section{Additional Results}
\label{sec:additional-results}

\subsection{Two- and Three-Expert Results}
\begin{table}[htbp]
\centering
\caption{SR (\%) after merging two or three experts. Avg is computed within each expert group. Continual methods follow WebShop $\rightarrow$ Tool $\rightarrow$ Search; batch methods merge the same expert subset in each group. Blue shading marks BACAM. \textbf{Bold} and \underline{underlined} values mark the highest and second-highest measured results per column; ties share the same mark.}
\label{tab:prefix-results}
\small
\begin{tabular}{@{}lrrrrrrr@{}}
\toprule
Method & \multicolumn{3}{c}{2 experts} & \multicolumn{4}{c}{3 experts} \\
\cmidrule(lr){2-4}\cmidrule(l){5-8}
 & WebShop & Tool & Avg & WebShop & Tool & Search & Avg \\
\midrule
\rowcolor{black!6}
\multicolumn{8}{l}{\textit{General model merging}} \\
Weight Average & 26.60 & \underline{1.00} & 13.80 & 7.00 & 0.00 & 22.21 & 9.74 \\
Task Arithmetic & 26.60 & \underline{1.00} & 13.80 & 24.60 & 1.00 & 21.86 & 15.82 \\
TIES-Merging & 31.20 & 0.00 & 15.60 & 30.60 & 0.00 & 16.31 & 15.64 \\
TSV-Merge & \underline{35.40} & \underline{1.00} & \underline{18.20} & \underline{36.80} & 0.00 & 23.75 & \underline{20.18} \\
WUDI-Merging & 34.40 & 0.00 & 17.20 & 33.60 & 0.00 & \textbf{24.84} & 19.48 \\
\midrule
\rowcolor{black!6}
\multicolumn{8}{l}{\textit{Agent model merging}} \\
RAM & 34.60 & 0.00 & 17.30 & 33.00 & 0.00 & 21.27 & 18.09 \\
RAM++ & 33.80 & 0.00 & 16.90 & 32.20 & \underline{2.00} & 21.17 & 18.46 \\
\midrule
\rowcolor{black!6}
\multicolumn{8}{l}{\textit{Continual model merging}} \\
OPCM & 33.20 & \underline{1.00} & 17.10 & 22.00 & 1.00 & 20.62 & 14.54 \\
NUFILT & 9.80 & 0.00 & 4.90 & 8.20 & 0.00 & 20.53 & 9.58 \\
\midrule
\rowcolor{blue!9}
\textbf{BACAM (ours)} & \textbf{78.60} & \textbf{32.00} & \textbf{55.30} & \textbf{78.40} & \textbf{37.00} & \underline{24.10} & \textbf{46.50} \\
\bottomrule
\end{tabular}
\end{table}

\subsection{Task Scores and Interaction Lengths}
Table~\ref{tab:interaction-results} complements the SR comparison with every reported TS and Round measurement. TS retains information about partial shopping-task completion, whereas SR requires complete success. Round includes successful and failed episodes, so comparisons should consider both task performance and interaction length. NUFILT results for WTSA and the earlier TSWA run are listed separately.

\begin{table}[htbp]
\centering
\caption{Task score and interaction length for all evaluated models. WebShop TS is on a 0--100 scale. Round averages interaction steps over all episodes, including failures. SR results appear in Table~\ref{tab:main-results}.}
\label{tab:interaction-results}
\small
\begin{tabular}{@{}lrrrr@{}}
\toprule
Method & \multicolumn{2}{c}{ALFWorld Round $\downarrow$} & \multicolumn{2}{c}{WebShop} \\
\cmidrule(lr){2-3}\cmidrule(lr){4-5}
 & IID & OOD & TS $\uparrow$ & Round $\downarrow$ \\
\midrule
\multicolumn{5}{l}{\textit{Base model and individual experts}} \\
Qwen2.5-7B-Instruct & 44.34 & 46.19 & 1.79 & 14.78 \\
Search expert & 48.00 & 47.10 & 4.45 & 14.49 \\
Tool expert & 47.32 & 47.19 & 1.96 & 14.85 \\
ALFWorld expert & 8.43 & 12.52 & 25.16 & 11.46 \\
WebShop expert & 42.02 & 42.78 & 88.41 & 6.11 \\
\midrule
\multicolumn{5}{l}{\textit{General model merging}} \\
Weight Average & 37.61 & 37.45 & 26.87 & 12.05 \\
Task Arithmetic & 24.19 & 28.20 & 56.66 & 9.79 \\
TIES-Merging & 28.64 & 29.00 & 62.00 & 9.52 \\
TA w/ DARE & 24.05 & 30.68 & 56.41 & 9.82 \\
TIES w/ DARE & 27.11 & 28.84 & 59.76 & 9.54 \\
TSV-Merge & 22.90 & 24.67 & 65.18 & 9.01 \\
Iso-C & 27.17 & 29.38 & 52.86 & 9.94 \\
WUDI-Merging & 21.46 & 24.31 & 54.24 & 10.01 \\
AdaMerging & 26.71 & 28.36 & 49.50 & 9.74 \\
AdaMerging++ & 26.20 & 26.20 & 53.15 & 9.98 \\
Expert Merging & 38.73 & 39.84 & 29.00 & 11.89 \\
\midrule
\multicolumn{5}{l}{\textit{Agent model merging}} \\
RAM & 29.09 & 30.15 & 61.54 & 9.06 \\
RAM++ & 27.48 & 31.49 & 61.09 & 9.31 \\
\midrule
\multicolumn{5}{l}{\textit{Continual model merging}} \\
OPCM & 32.69 & 35.47 & 44.76 & 10.76 \\
NUFILT & 37.49 & 39.31 & 32.73 & 11.79 \\
NUFILT (TSWA reference) & 37.50 & 36.94 & 28.83 & 11.95 \\
\midrule
BACAM (ours) & 14.01 & 14.69 & 88.16 & 6.09 \\
\bottomrule
\end{tabular}
\end{table}

\FloatBarrier

\subsection{Stage-wise Scores and Capability Acquisition}
\label{sec:stage-plots}

Table~\ref{tab:stage-results} reports the scores underlying Figure~\ref{fig:merging-stages}, together with WebShop TS.

\begin{table}[htbp]
\centering
\caption{BACAM after each integration for the measured orders. All task columns report SR (\%) except the final WebShop TS column. -- marks a task whose expert has not yet been integrated.}
\label{tab:stage-results}
\small
\begin{tabular}{@{}llrrrrrr@{}}
\toprule
Order & Stage & WebShop & Tool & Search & \multicolumn{2}{c}{ALFWorld} & WebShop \\
\cmidrule(lr){6-7}
 & & SR & SR & SR & IID SR & OOD SR & TS \\
\midrule
WTSA & 1: W & 78.00 & -- & -- & -- & -- & 88.41 \\
 & 2: + T & 78.60 & 32.00 & -- & -- & -- & 88.58 \\
 & 3: + S & 78.40 & 37.00 & 24.10 & -- & -- & 88.32 \\
 & 4: + A & 77.20 & 38.00 & 23.75 & 88.57 & 86.57 & 88.16 \\
\midrule
ASTW & 1: A & -- & -- & -- & 98.57 & 91.04 & -- \\
 & 2: + S & -- & -- & 24.14 & 96.43 & 91.04 & -- \\
 & 3: + T & -- & 31.00 & 22.56 & 97.14 & 90.30 & -- \\
 & 4: + W & 63.00 & 27.00 & 24.05 & 96.43 & 91.04 & 77.51 \\
\midrule
TWAS & 1: T & -- & 30.00 & -- & -- & -- & -- \\
 & 2: + W & 67.80 & 30.00 & -- & -- & -- & 80.95 \\
 & 3: + A & 64.20 & 23.00 & -- & 88.57 & 82.84 & 79.09 \\
 & 4: + S & 56.80 & 38.00 & 25.43 & 84.29 & 78.36 & 71.39 \\
\midrule
SAWT & 1: S & -- & -- & 24.34 & -- & -- & -- \\
 & 2: + A & -- & -- & 24.44 & 90.00 & 83.58 & -- \\
 & 3: + W & 65.60 & -- & 24.74 & 88.57 & 82.09 & 79.09 \\
 & 4: + T & 58.20 & 33.00 & 25.43 & 82.86 & 80.60 & 73.69 \\
\bottomrule
\end{tabular}
\end{table}


\FloatBarrier
\begin{table}[htbp]
\centering
\caption{Stage-wise SR (\%) on Llama-3.1-8B-Instruct under WTSA.}
\label{tab:llama-stage-results}
\footnotesize
\begin{tabular}{@{}llrrrrr@{}}
\toprule
Stage & Incoming task & WebShop & Tool & Search & ALFWorld IID & ALFWorld OOD \\
\midrule
T2 & Tool & 61.20 & 42.00 & -- & -- & -- \\
T3 & Search & 62.00 & 40.00 & 22.71 & -- & -- \\
T4 & ALFWorld & 62.40 & 41.00 & 21.67 & 62.86 & 59.70 \\
\bottomrule
\end{tabular}
\end{table}

\FloatBarrier
\subsection{Sensitivity to Merging Order}

We attribute order sensitivity to the differences in new-task capability acquisition across tasks, which is not caused by gate saturation, Lagrangian-multiplier dynamics, the interaction between token-group weighting and relatively sparse action tokens. Our method remains robust across the merging orders. The order determines the initial expert and which capabilities must be acquired later. For example, in WebShop $\rightarrow$ Tool, WebShop is the initial expert and Tool is the new task. In the reverse order, Tool is the initial expert and requires acquiring WebShop. Table~\ref{tab:order-acquisition} shows lower acquisition relative to the expert for WebShop (83.93\%) and ALFWorld (90.34\% IID and 92.63\% OOD) than for Search (100.90\%) and Tool (106.67\%), which is caused by differences in trajectory coverage and how clearly teacher supervision distinguishes between decisions across tasks. Within each task, performance at the integration stage shows limited variation across merging orders, with a maximum standard deviation of about 1.96 percentage points. Table~\ref{tab:order-forgetting} shows limited forgetting across the merging orders. Together, these results suggest that order sensitivity mainly concerns the differences in new-task capability acquisition across tasks, while capability acquisition within each task and retention of existing capabilities remain robust across merging orders.

\begin{table}[htbp]
\centering
\caption{New-task capability acquisition across merging orders.}
\label{tab:order-acquisition}
\footnotesize
\setlength{\tabcolsep}{3pt}
\renewcommand{\arraystretch}{1.1}
\begin{tabular}{@{}lrrrrr@{}}
\toprule
Merging order & WebShop & Tool & Search & ALFWorld IID & ALFWorld OOD \\
\midrule
WTSA & Initial expert & 32.00 (T2) & 24.10 (T3) & 88.57 (T4) & 86.57 (T4) \\
ASTW & 63.00 (T4) & 31.00 (T3) & 24.14 (T2) & Initial expert & Initial expert \\
TWAS & 67.80 (T2) & Initial expert & 25.43 (T4) & 88.57 (T3) & 82.84 (T3) \\
SAWT & 65.60 (T3) & 33.00 (T4) & Initial expert & 90.00 (T2) & 83.58 (T2) \\
Mean $\pm$ SD & 65.47 $\pm$ 1.96 & 32.00 $\pm$ 0.82 & 24.56 $\pm$ 0.62 & 89.05 $\pm$ 0.67 & 84.33 $\pm$ 1.61 \\
Capability acquisition rate & 83.93\% & 106.67\% & 100.90\% & 90.34\% & 92.63\% \\
\bottomrule
\end{tabular}
\end{table}

\noindent\textit{Note.} Tn denotes the stage at which a task is introduced. Capability acquisition measures performance on a newly integrated task relative to its expert. The expert reference scores are 78.00, 30.00, 24.34, 98.57, and 91.04, respectively.

\begin{table}[htbp]
\centering
\caption{Forgetting.}
\label{tab:order-forgetting}
\footnotesize
\setlength{\tabcolsep}{3pt}
\renewcommand{\arraystretch}{1.1}
\begin{tabular}{@{}lrrrrr@{}}
\toprule
Merging order & Initial task & Expert score & Final score & Drop (pp) & Retention \\
\midrule
WTSA & WebShop & 78.00 & 77.20 & 0.80 & 98.97\% \\
ASTW & ALFWorld IID & 98.57 & 96.43 & 2.14 & 97.83\% \\
ASTW & ALFWorld OOD & 91.04 & 91.04 & 0.00 & 100.00\% \\
TWAS & Tool & 30.00 & 38.00 & $-$8.00 & 126.67\% \\
SAWT & Search & 24.34 & 25.43 & $-$1.09 & 104.49\% \\
\bottomrule
\end{tabular}
\end{table}

\noindent\textit{Note.} Each order reports forgetting only for its initial task.

\FloatBarrier
\subsection{Generalization to Longer Merging Sequences}

We extend the four-task setting with the Airline task from $\tau^2$-bench to evaluate a longer merging sequence (Table~\ref{tab:longer-merging-sequence}). The five-stage order is WebShop, Tool, Search, ALFWorld, and $\tau^2$-bench. The results show limited performance loss on old tasks and effective acquisition of the new task.

\begin{table}[htbp]
\centering
\caption{Performance over a longer merging sequence.}
\label{tab:longer-merging-sequence}
\footnotesize
\setlength{\tabcolsep}{3pt}
\renewcommand{\arraystretch}{1.1}
\begin{tabular}{@{}lrrrrrr@{}}
\toprule
Method & WebShop & Tool & Search & ALFWorld IID & ALFWorld OOD & $\tau^2$-bench \\
\midrule
BACAM few-shot T4 (ours) & 76.80 & 28.00 & 23.15 & 86.43 & 79.10 & - \\
BACAM few-shot T5 (ours) & 75.40 & 25.00 & 21.72 & 79.29 & 74.63 & 40.91 \\
Forgetting & 1.40 & 3.00 & 1.44 & 7.14 & 4.48 & - \\
\bottomrule
\end{tabular}
\end{table}

\begin{enumerate}
\item \textbf{Benchmark.} We use the Airline task from $\tau^2$-bench~\citep{barres2025tau2bench}.
\item \textbf{Evaluation framework.} We use the $\tau^2$-bench implementation provided by Simia~\citep{li2025simia}. Both the user simulator and the natural-language judge use the qwen3.8-max API.
\item \textbf{Expert checkpoint.} We use Simia-Tau-SFT-RL-Qwen2.5-7B.\footnote{\url{https://huggingface.co/Simia-Agent/Simia-Tau-SFT-RL-Qwen2.5-7B}} Its Airline success rate under these settings is 47.73\%.
\end{enumerate}

\FloatBarrier
\section{Gate and Conflict Diagnostics}
\label{sec:additional-analysis}

\subsection{TWAS Order}
\label{sec:gate-diagnostics}

Under TWAS, gate movement declines across successive integrations, but its allocation across tensors remains structured. MLP tensors account for most movement, while attention gains share as the sequence progresses. Plasticity budgets vary more across capabilities and stages than gate movement does, indicating that the controller adapts protection to the incoming capability.

\begin{figure}[htbp]
\centering
\includegraphics[width=\linewidth]{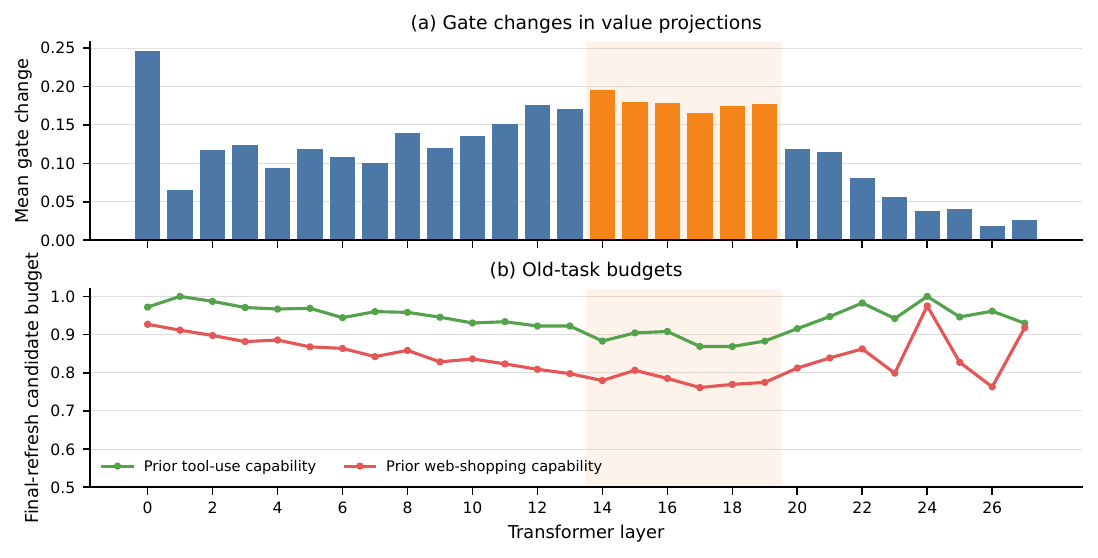}
\caption{Value-projection tensors during ALFWorld integration in TWAS. (a) Mean gate change by layer, with layers 14--19 highlighted. (b) Final-refresh candidate budgets for Tool and WebShop; lower budgets indicate stronger protection. WebShop is more tightly protected across the highlighted layers.}
\label{fig:gate-value-conflict}
\end{figure}

\begin{figure}[htbp]
\centering
\includegraphics[width=\linewidth]{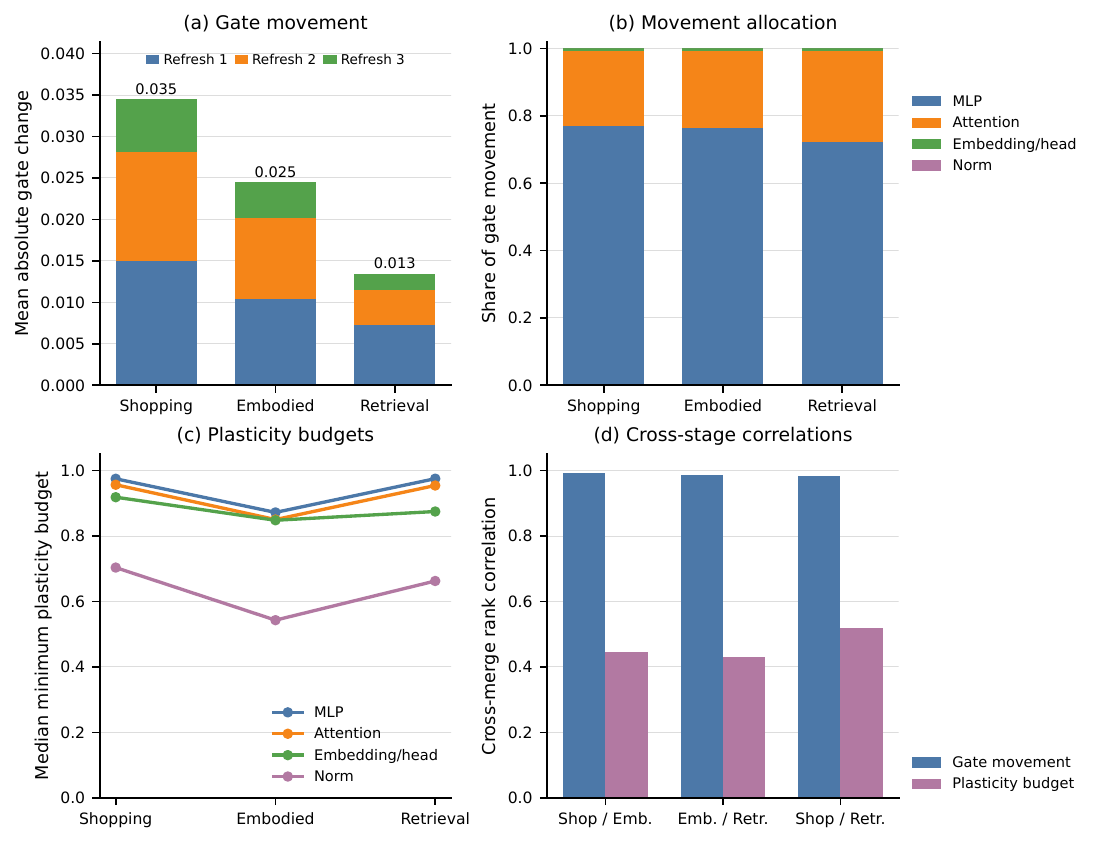}
\caption{Gate movement and plasticity budgets across TWAS integrations, ordered by incoming WebShop, ALFWorld, and Search experts. (a) Per-refresh mean absolute gate change, which declines across stages; (b) movement share by tensor family, dominated by MLP; (c) median minimum budget by family, where lower values mean stronger protection; and (d) cross-stage Spearman rank correlations, showing more persistent movement than budgets.}
\label{fig:gate-diagnostics-overview}
\end{figure}

\begin{figure}[htbp]
\centering
\includegraphics[width=\linewidth]{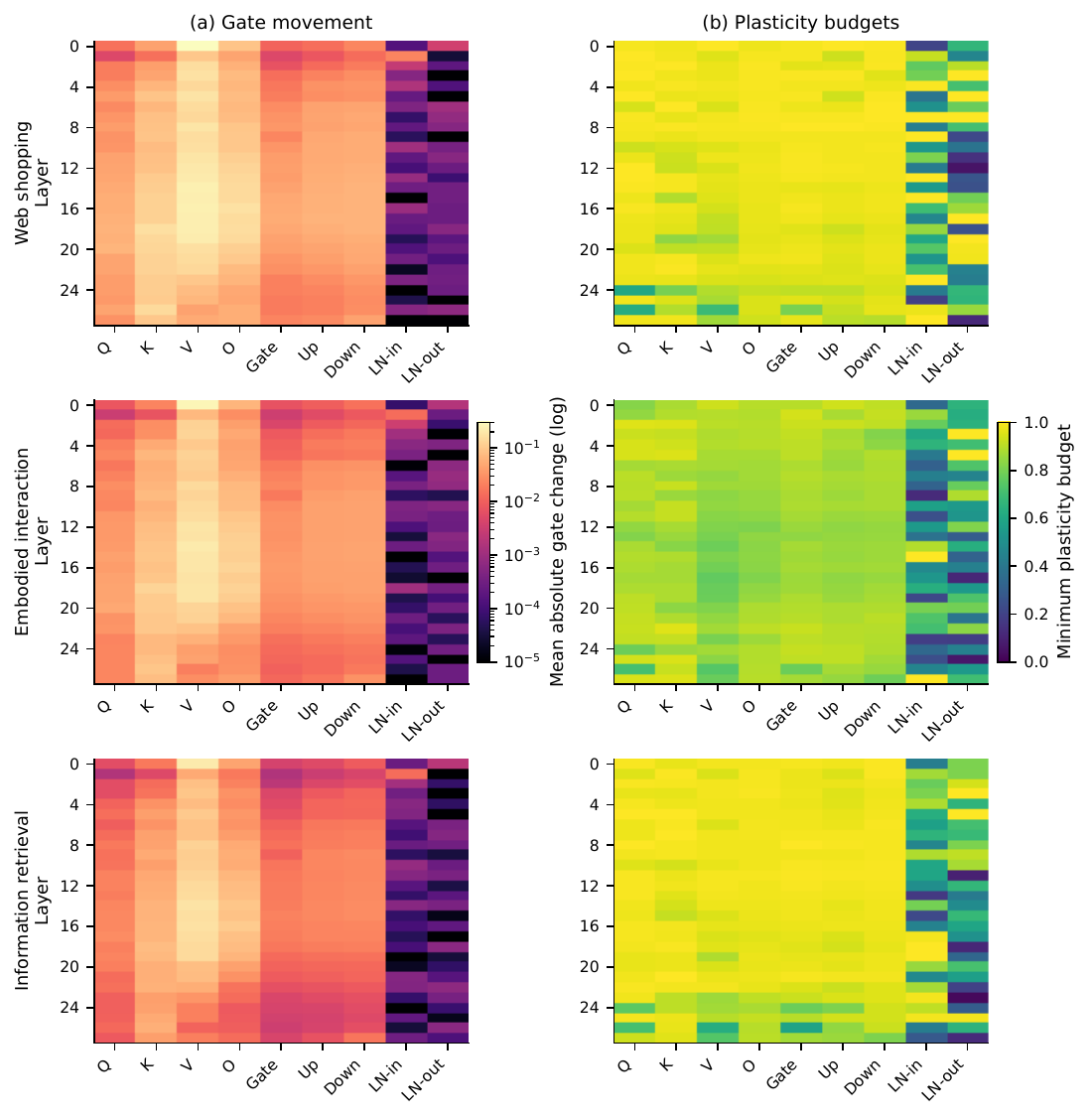}
\caption{Layer-wise diagnostics across TWAS integrations. Rows follow incoming WebShop, ALFWorld, and Search experts; columns show (a) mean absolute gate movement and (b) minimum plasticity budget by layer and weight type. Lower budgets indicate stronger protection, concentrated in selected tensors.}
\label{fig:gate-diagnostics-layer-map}
\end{figure}

\begin{figure}[htbp]
\centering
\includegraphics[width=\linewidth]{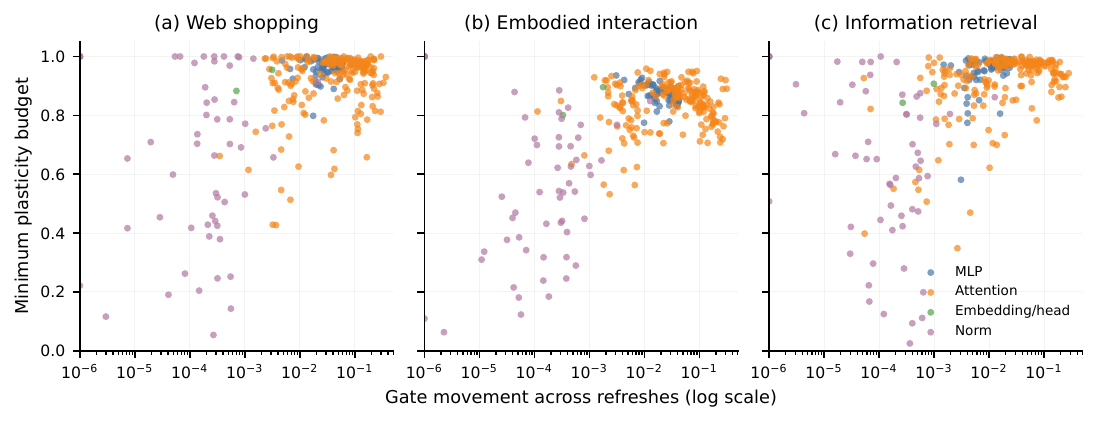}
\caption{Tensor-level gate movement versus minimum plasticity budget across TWAS integrations. Each point is one tensor; (a)--(c) correspond to incoming WebShop, ALFWorld, and Search experts. Lower budgets indicate stronger protection.}
\label{fig:gate-diagnostics-scatter}
\end{figure}
\FloatBarrier

\subsection{SAWT Order}
The same tensor-level diagnostics are reported for the Search--ALFWorld--WebShop--Tool order.

\begin{figure}[htbp]
\centering
\includegraphics[width=\linewidth]{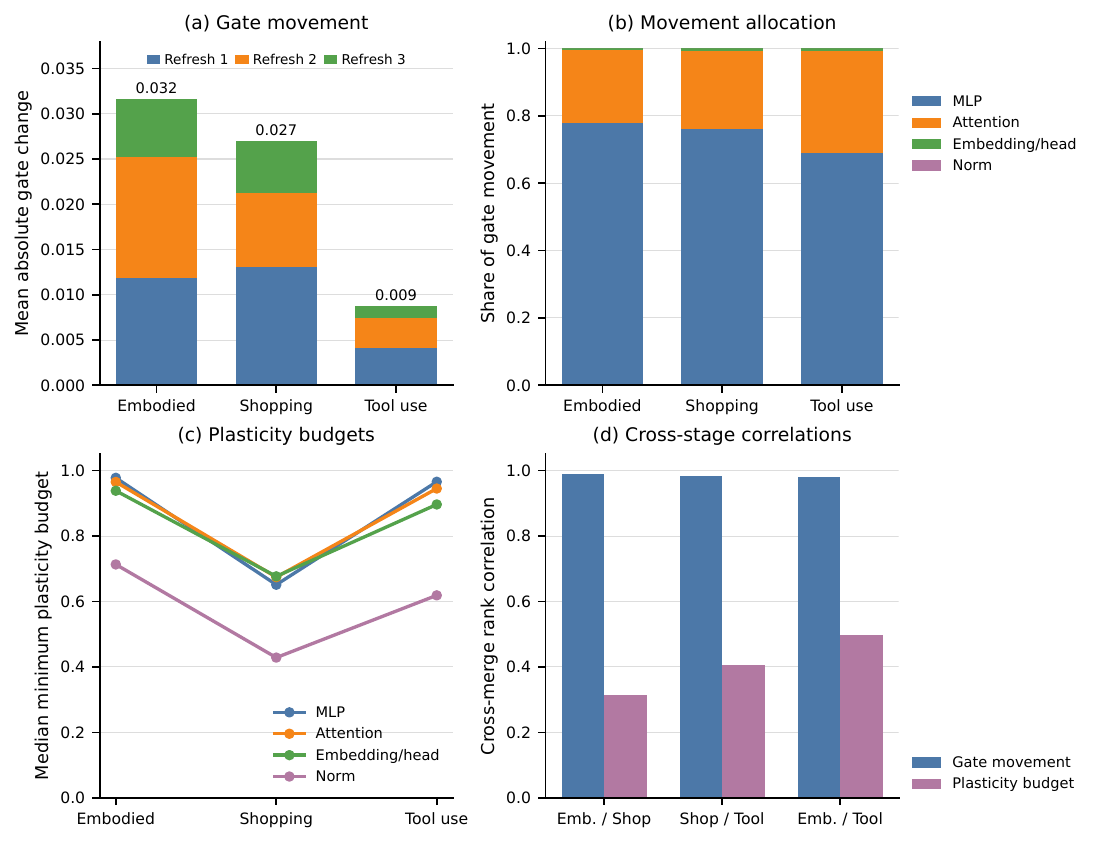}
\caption{Gate movement and plasticity budgets across SAWT integrations, ordered by incoming ALFWorld, WebShop, and Tool experts. (a) Per-refresh mean absolute gate change, which declines across stages; (b) movement share by tensor family, dominated by MLP; (c) median minimum budget by family, where lower values mean stronger protection; and (d) cross-stage Spearman rank correlations, showing more persistent movement than budgets.}
\label{fig:gate-diagnostics-overview-sawt}
\end{figure}

\begin{figure}[htbp]
\centering
\includegraphics[width=\linewidth]{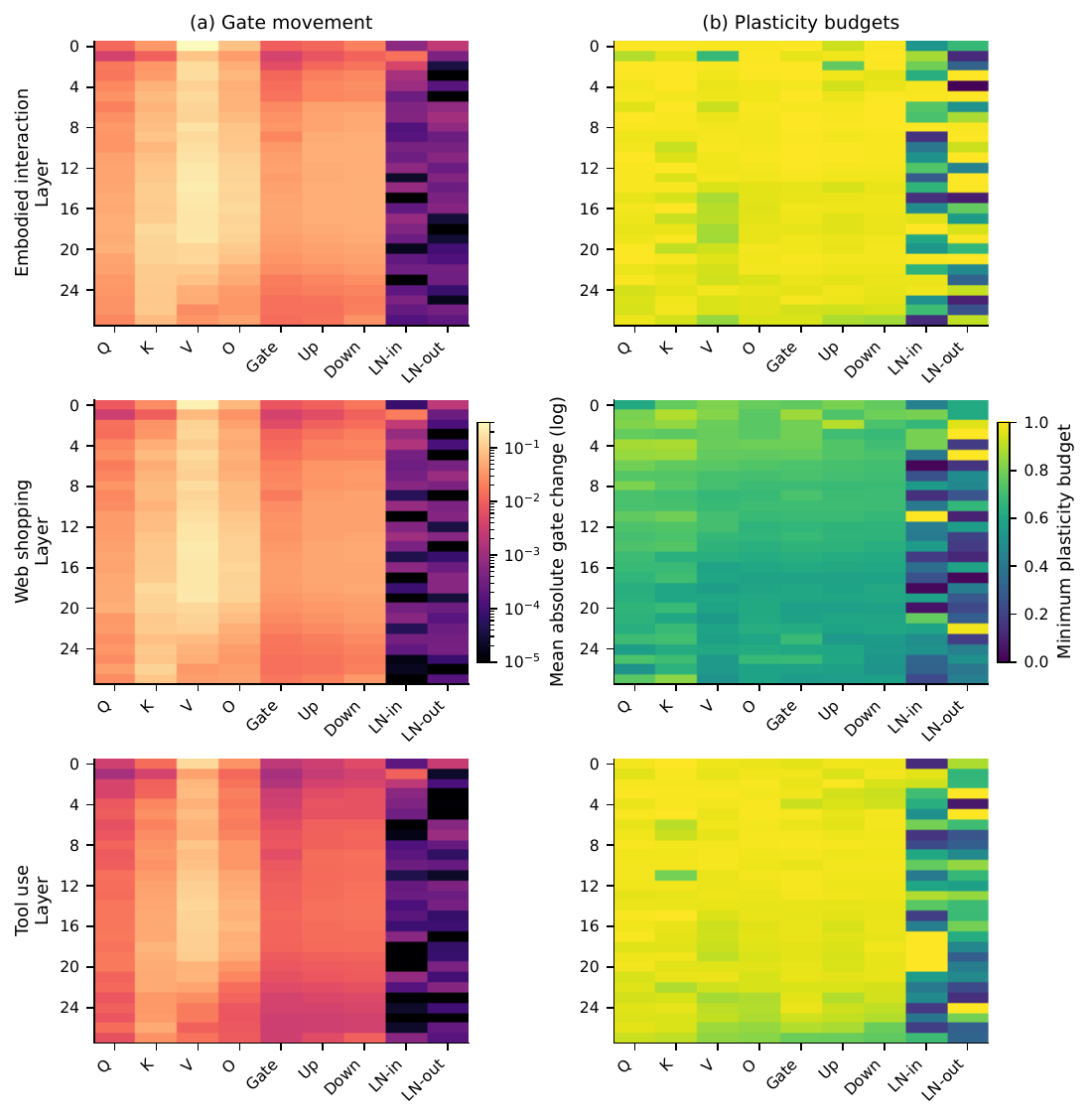}
\caption{Layer-wise diagnostics across SAWT integrations. Rows follow incoming ALFWorld, WebShop, and Tool experts; columns show (a) mean absolute gate movement and (b) minimum plasticity budget by layer and weight type. Lower budgets indicate stronger protection.}
\label{fig:gate-diagnostics-layer-map-sawt}
\end{figure}

\begin{figure}[htbp]
\centering
\includegraphics[width=\linewidth]{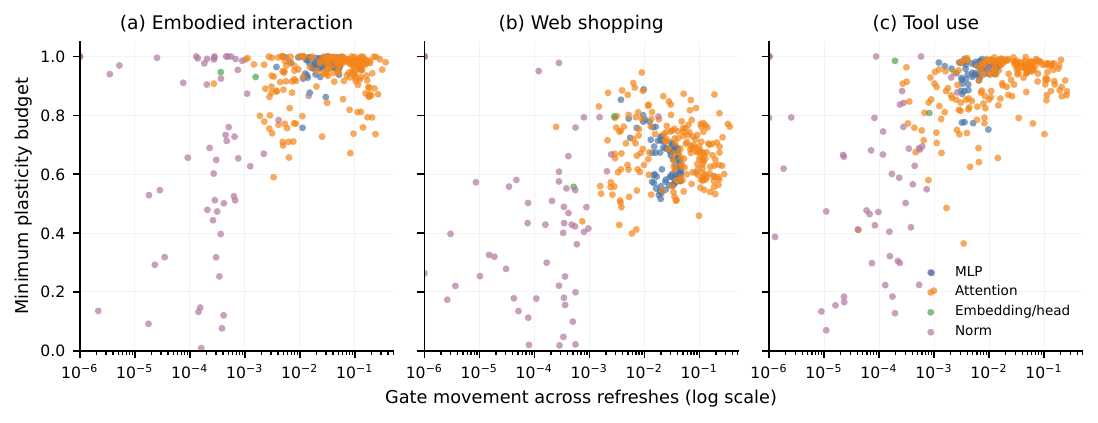}
\caption{Tensor-level gate movement versus minimum plasticity budget across SAWT integrations. Each point is one tensor; (a)--(c) correspond to incoming ALFWorld, WebShop, and Tool experts. Lower budgets indicate stronger protection.}
\label{fig:gate-diagnostics-scatter-sawt}
\end{figure}

\begin{figure}[htbp]
\centering
\includegraphics[width=\linewidth]{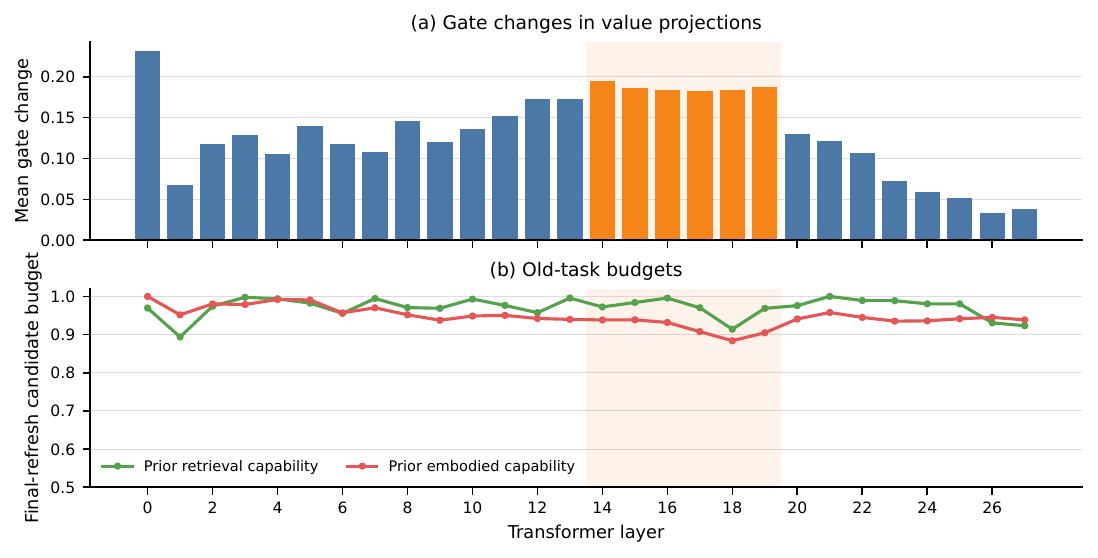}
\caption{Value-projection tensors during WebShop integration in SAWT. (a) Mean gate change by layer, with layers 14--19 highlighted. (b) Final-refresh candidate budgets for Search and ALFWorld; lower budgets indicate stronger protection. ALFWorld is more tightly protected across the highlighted layers.}
\label{fig:gate-value-conflict-sawt}
\end{figure}
\FloatBarrier

\clearpage
\section{Qualitative Examples}
\label{sec:qualitative-examples}

\newtcolorbox{bacamcase}[1]{colback=blue!3!white,colframe=blue!35!black,
  title={#1},fonttitle=\bfseries,boxrule=0.5pt,arc=1mm,
  left=5pt,right=5pt,top=4pt,bottom=4pt}
\newtcolorbox{bacamcorrect}{colback=green!6!white,colframe=green!45!black,
  title={BACAM: Correct},fonttitle=\bfseries,boxrule=0.5pt,arc=1mm,
  left=5pt,right=5pt,top=4pt,bottom=4pt}
\newtcolorbox{bacamincorrect}[1]{colback=red!6!white,colframe=red!45!black,
  title={#1: Incorrect},fonttitle=\bfseries,boxrule=0.5pt,arc=1mm,
  left=5pt,right=5pt,top=4pt,bottom=4pt}

The examples compare BACAM with WUDI-Merging on matched evaluation instances. Each example summarizes the decisive observation--reasoning--action step; reasoning is paraphrased from the recorded trajectory rather than quoted verbatim.

\begingroup\small
\begin{bacamcase}{WebShop: tracking a remaining product constraint}
\textbf{Observation:} The requested curtain is open. The color ``sea mist'' has been selected, but the required 52-by-84-inch size remains to be selected.
\end{bacamcase}
\noindent\begin{minipage}[t]{0.49\linewidth}
\vspace{0pt}
\begin{bacamcorrect}
\textbf{Reasoning:} The size constraint is still unmet.\par
\textbf{Action:} Select 52-by-84-inch size, then purchase.\par
\textbf{Outcome:} Success; task score 1.00.
\end{bacamcorrect}
\end{minipage}\hfill
\begin{minipage}[t]{0.49\linewidth}
\vspace{0pt}
\begin{bacamincorrect}{WUDI-Merging}
\textbf{Reasoning:} Select the requested color.\par
\textbf{Action:} Select ``sea mist'' again, then purchase.\par
\textbf{Outcome:} Size omitted; task score 0.80.
\end{bacamincorrect}
\end{minipage}

\smallskip
\begin{bacamcase}{Tool use: selecting the function that matches the request}
\textbf{Observation:} The user asks for the latest trading details of Quasar Ltd.
\end{bacamcase}
\noindent\begin{minipage}[t]{0.49\linewidth}
\vspace{0pt}
\begin{bacamcorrect}
\textbf{Reasoning:} Resolve the company's ticker, then retrieve its stock information.\par
\textbf{Action:} \texttt{get\_symbol\_by\_name}, then \texttt{get\_stock\_info}.\par
\textbf{Outcome:} Correct under the BFCL evaluator.
\end{bacamcorrect}
\end{minipage}\hfill
\begin{minipage}[t]{0.49\linewidth}
\vspace{0pt}
\begin{bacamincorrect}{WUDI-Merging}
\textbf{Reasoning:} Treat the request as asking for past transactions.\par
\textbf{Action:} Call \texttt{get\_transaction\_history}.\par
\textbf{Outcome:} Tool response does not match the request.
\end{bacamincorrect}
\end{minipage}

\smallskip
\begin{bacamcase}{Search: resolving a multi-hop clue before answering}
\textbf{Observation:} The question asks for the largest national park in the country identified by a hunting-versus-ecotourism spending clue.
\end{bacamcase}
\noindent\begin{minipage}[t]{0.49\linewidth}
\vspace{0pt}
\begin{bacamcorrect}
\textbf{Reasoning:} Resolve the country clue before searching for its largest park.\par
\textbf{Action:} Search for Tanzania's largest national park.\par
\textbf{Outcome:} Ruaha National Park (correct).
\end{bacamcorrect}
\end{minipage}\hfill
\begin{minipage}[t]{0.49\linewidth}
\vspace{0pt}
\begin{bacamincorrect}{WUDI-Merging}
\textbf{Reasoning:} Assume the country is the United States.\par
\textbf{Action:} Search for the largest US national park.\par
\textbf{Outcome:} Yellowstone National Park (incorrect).
\end{bacamincorrect}
\end{minipage}

\smallskip
\begin{bacamcase}{ALFWorld: responding to the latest environment state}
\textbf{Observation:} The agent has picked up the bowl from the desk and must examine it with a desk lamp.
\end{bacamcase}
\noindent\begin{minipage}[t]{0.49\linewidth}
\vspace{0pt}
\begin{bacamcorrect}
\textbf{Reasoning:} The lamp is not at the desk; check another location.\par
\textbf{Action:} Go to the dresser, then use the lamp.\par
\textbf{Outcome:} Success in 4 steps.
\end{bacamcorrect}
\end{minipage}\hfill
\begin{minipage}[t]{0.49\linewidth}
\vspace{0pt}
\begin{bacamincorrect}{WUDI-Merging}
\textbf{Reasoning:} Look for the lamp after picking up the bowl.\par
\textbf{Action:} Return to the desk, then repeat invalid lamp commands.\par
\textbf{Outcome:} No success within 50 steps.
\end{bacamincorrect}
\end{minipage}
\endgroup

Across these matched examples, BACAM follows the remaining task constraints and environment feedback at the decisive step, whereas WUDI-Merging diverges before completion.

\end{document}